\documentclass[10pt,twocolumn,letterpaper]{article}

\usepackage{iccv}
\pdfoutput=1
\makeatletter
\@namedef{ver@everyshi.sty}{}
\makeatother
\usepackage{tikz}

\usepackage{times}
\usepackage{epsfig}
\usepackage{graphicx}
\usepackage{amsmath}
\usepackage{amssymb}

\usepackage{booktabs}

\usepackage{makecell}
\usepackage{multirow}
\usepackage{titling}

\usepackage{MnSymbol}
\usepackage{comment}

\newcommand{\bx}{\boldsymbol{x}}

\newcommand{\by}{\boldsymbol{y}}
\newcommand{\bY}{\boldsymbol{Y}}

\usepackage{xcolor,colortbl}

\definecolor{Gray}{gray}{0.85}
\definecolor{LightCyan}{rgb}{0.88,1,1}
\newcolumntype{a}{>{\columncolor{Gray}}c}
\newcolumntype{b}{>{\columncolor{white}}c}
\usepackage{nicematrix}

\usepackage[pagebackref=true,breaklinks=true,letterpaper=true,colorlinks,bookmarks=false]{hyperref}

\iccvfinalcopy 

\def\iccvPaperID{****} 

\ificcvfinal\fi

\begin{document}

\title{\vspace{-1.02cm}
\textbf{Fairness meets Cross-Domain Learning: \\
a new perspective on Models and Metrics}}

\author{
Leonardo Iurada$^{1}$ \hspace{0.5cm} Silvia Bucci$^{1}$\thanks{Work mainly developed during the internship period at  the University of Edinburgh.} \hspace{0.5cm} Timothy M. Hospedales$^{3}$ \hspace{0.5cm} Tatiana Tommasi$^{1,2}$\\ 
$^{1}$Politecnico di Torino  \hspace{0.5cm} $^{2}$Italian Institute of Technology \hspace{0.5cm} $^{3}$University of Edinburgh \\
\texttt{\small \{leonardo.iurada, silvia.bucci, tatiana.tommasi\}@polito.it} \hspace{0.5cm} \texttt{\small t.hospedales@ed.ac.uk} 
}

\date{}

\maketitle

\ificcvfinal\thispagestyle{empty}\fi

\begin{abstract}
    Deep learning-based recognition systems are deployed at scale for several real-world applications that inevitably involve our social life. Although being of great support when making complex decisions, they might capture spurious data correlations and leverage sensitive attributes (e.g. age, gender, ethnicity). How to factor out this information while keeping a high prediction performance is a task with still several open questions, many of which are shared with those of the domain adaptation and generalization literature which focuses on avoiding visual domain biases. 

In this work, we propose an in-depth study of the relationship between
cross-domain learning (CD) and model fairness by introducing a benchmark on face and medical images spanning several demographic groups as well as classification and localization tasks.
After having highlighted the limits of the current evaluation metrics, we introduce a new \emph{Harmonic Fairness (HF)} score to assess jointly how fair and accurate every model is with respect to a reference baseline. 
Our study covers 14 CD approaches alongside three state-of-the-art fairness algorithms and shows how the former can outperform the latter. 
Overall, our work paves the way for a more systematic analysis of fairness problems in computer vision. \emph{Code available at: \url{https://github.com/iurada/fairness_crossdomain}}

\end{abstract}

\section{Introduction}

Deep neural networks currently constitute the core of several AI systems that support decisions in many socially important tasks like the hiring process, healthcare diagnosis, and law enforcement. Despite their efficacy in aggregate, it has become apparent that they 
can learn to encode subtle biases that disproportionately disadvantage particular sub-populations (\eg based on age, gender, ethnicity, etc.) \cite{buolamwini2018gender,berk2021fairness,seyyed2021underdiagnosis,obermeyer2019dissecting}. The causes of this unfairness
are many, from amplifying bias that already exists in the training data \cite{wang2020towards}, to learning spurious correlations \cite{geirhos2020shortcut}. However, the end result is the same: AI systems may exacerbate rather than alleviate social problems of inequality and discrimination. This issue has motivated a growing body of research in fairness interventions \cite{wang2020towards,zietlow2022leveling,park2022fair} — algorithms that are particularly designed to optimize some notion of fairness simultaneously with conventional learning objectives. Overall, the research in this area is still in its infancy and several factors have been overlooked till now. 

\begin{figure}[t]
\centering
\includegraphics[width=0.95\linewidth]{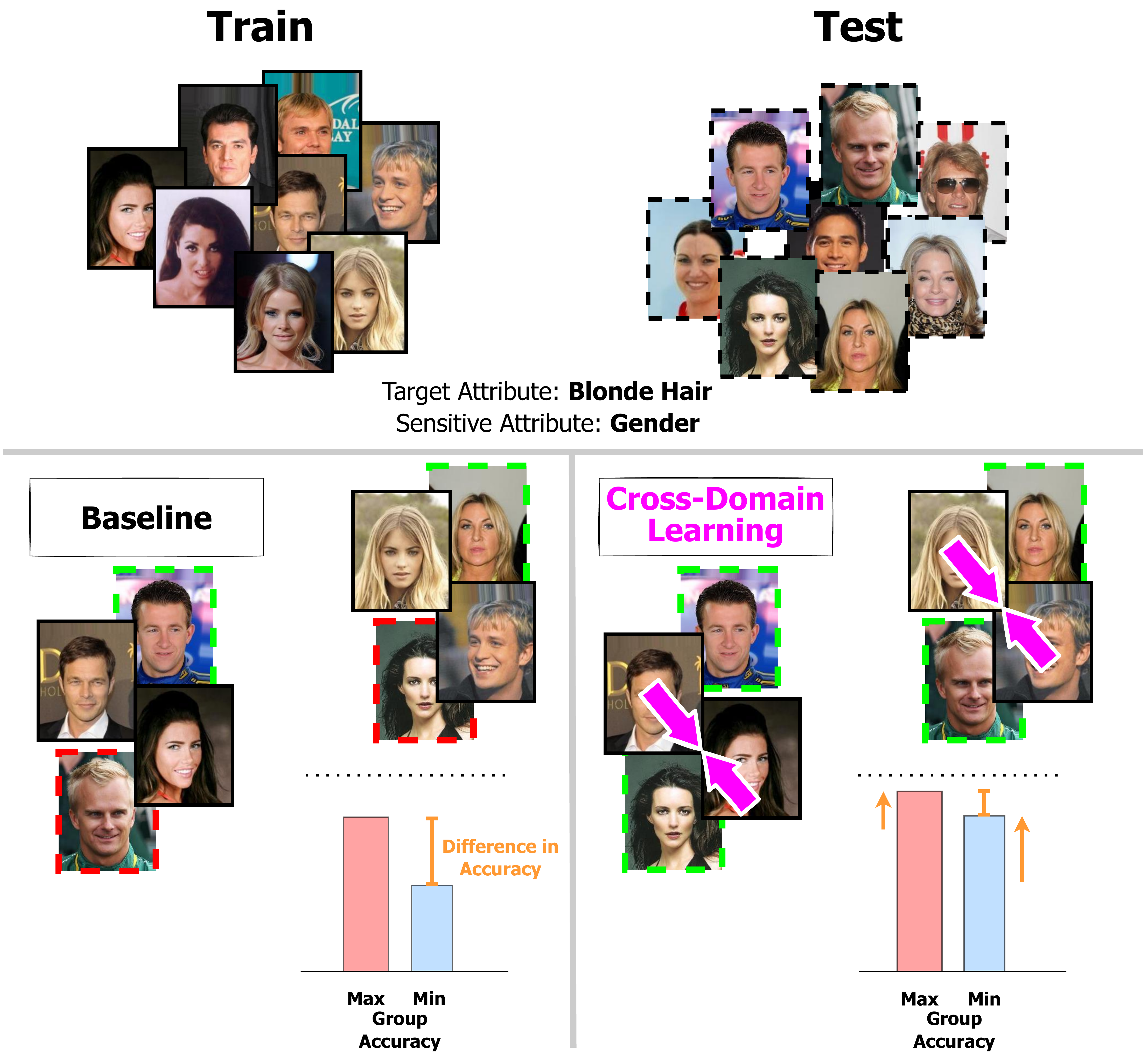}
\caption{
We start from a model that exhibits some degree of unfairness as evidenced by the Difference in Accuracy between protected groups (left). We exploit Cross-Domain (CD) learning to reduce the visual domain shift among groups and improve the generalization ability of the model, obtaining an unfairness mitigation effect (right).
}
\label{fig:teaser} \vspace{-5mm}
\end{figure}

One is the natural alignment of the fair learning problem with the more widely studied cross-domain (CD) learning challenge in computer vision. In the latter area, the goal is to produce models agnostic to the specific details of visual domains (\eg camera pose, lighting, image style)
to get generalization across them. 
By mapping visual domains to protected subgroups, we can see that the wealth of existing algorithms for promoting domain invariance could potentially benefit fairness (see Figure \ref{fig:teaser}). Thus, \textbf{our first contribution is to introduce a new fairness benchmark for computer vision}. It spans both face 
and medical images for classification and landmark detection tasks  
and compares 14 CD learning approaches alongside three state-of-the-art (SOTA) fairness algorithms.
We remark that current unfairness mitigation strategies in computer vision are restricted to classification problems. To overcome this limitation we include in our benchmark the task of landmark detection on face images of different demographic groups, as the bias related to sensitive attributes can affect the precision with which critical keypoints are located.

Another aspect on which there is still a lot of confusion and open debate is about how systems should be evaluated. There are multiple competing notions of fairness and ways to quantify it \cite{verma2018fairnessExplained}. 
Previous studies measure group fairness by accuracy difference between advantaged and disadvantaged subgroups \cite{hardt2016equality}. However, this goal has been criticized in philosophy and ethics literature \cite{mason2001egalitarianism}. Purely minimizing the gap between subgroup performance, may lead to choosing a model with worse accuracy for all subgroups, which is Pareto inefficient \cite{zietlow2022leveling} and violates the ethical principles of beneficence and non-maleficence \cite{beauchamp2003methods}. 
\textbf{As our second contribution, we analyze existing group fairness criteria and propose a novel evaluation metric named \emph{Harmonic Fairness} that properly aggregates overall performance and fairness level to assess the quality of a model.} After having explained its design process, we use it as the basis for all the evaluations of our benchmark. 

The results of our extensive experimental analysis confirm the effectiveness of CD methods and the relevance of the proposed metric. It highlights how less popular approaches in the CD literature provide a significant advantage for unfairness mitigation on different tasks, systematically outperforming the tailored SOTA approaches. Moreover, it  shows that CD models trained to overcome the bias due to one sensitive attribute can be beneficial also to prevent unfairness with respect to a different one. This transfer ability provides insights into the robustness of CD approaches for fairness applications.
Overall, our work paves the way for a more systematic analysis of fairness problems in computer vision and the related unfairness mitigation methods, providing reliable tools for future evaluations.

\section{Related Works}
\noindent\textbf{Mitigating Unfairness.} 
The concept of fairness is very broad and has been largely discussed in the machine learning literature 
to support social, economic, and law choices  \cite{brennan2013emergence,van2019hiring,berk2019accuracy,jain2017weapons}. 
For our work, we focus on group fairness whose aim is to develop decision techniques 
that are invariant to differences across non-overlapping subsets of data defined by human-sensitive attributes like gender and ethnicity. 
Several studies have been conducted on face and medical image collections
to demonstrate how their biases lead to poor performing recognition models on some minority groups, progressively attracting the attention of the computer vision community \cite{kleinberg2018discrimination,zong2023medfair}. 
The existing strategies developed to mitigate unfairness have tackled the problem at three main levels depending on when they are applied within the learning process. 
As data unbalancing is among the main sources of unfairness, some methods act \emph{before training} by
collecting ad hoc datasets \cite{Kark2021fairface}, introducing strategic sampling  \cite{wang2020towards} or developing generative models that mitigate the imbalance through image synthesis \cite{zietlow2022leveling,kortylewski2019analyzing, wang2020towards,sattigeri2019fairness}. 
Other techniques have been designed to prevent models from capturing spurious data correlations \emph{during training}, by improving the 
representation learning procedure. Some approaches quantify these correlations and minimize them by aligning the representations of different demographic groups \cite{jung2021fair,wang2019racial}. 
Disentanglement-based approaches force orthogonality between target classes and sensitive attributes in order to disregard the latter during task learning \cite{dwork2018decoupled,lee2021learning,tartaglione2021end,sarhan2020fairness}. A similar goal is obtained by adversarial approaches that include dedicated modules to reduce the discriminability of semantic attributes \cite{dhar2021pass,kehrenberg2020null,diana2021minimax,gong2020jointly,wang2020towards}. 
Other approaches leverage feature distillation \cite{jung2021fair}, reinforcement \cite{wang2020mitigating} and contrastive learning \cite{park2022fair}. Very recently, a different family of methods proposed to identify and remove the critical parts of the models causing unfairness \cite{zhang2022recover,Savani2020intra}.
Finally, \emph{post-processing} techniques modify output predictions on the basis of fairness criteria \cite{Kim2019multiaccuracy}.


\vspace{1mm}\noindent\textbf{Cross-Domain Learning.} 
In real-world conditions training and test data often belong to different domains. Cross-domain models are trained to provide good performance on any unseen target domain at test time (\emph{Domain Generalization}), or to adapt the training source knowledge to a specific, different but related target (\emph{Domain Adaptation}). 

The techniques proposed to tackle the challenging \textit{Single-Source Domain Generalization} (SSDG) setting extend regularization strategies usually applied in empirical risk minimization to prevent overfitting (\eg label smoothing \cite{szegedy2016rethinking}), and reshape them to face large source-target domain shifts. 
These include strategic dropout based on gradient observation \cite{huang2020self}, tailored model selection \cite{cha2021swad,izmailov2018averaging} or data-augmentation  to increase data variability \cite{wang2021learning}. 
%
When training samples are drawn from multiple domains,
robust models can be obtained via data-augmentation techniques 
\cite{xu2020adversarial}, or style-transfer-based approaches \cite{borlino2021rethinking}.
Other popular \textit{Multi-Source Domain Generalization} (MSDG) strategies 
align the source domain representations through Maximum-Mean Discrepancy (MMD) minimization \cite{li2018domain} or adversarial learning \cite{li2018deep,kim2021selfreg}. 
A similar aim is also pursued by 
multi-task models that combine supervised and self-supervised learning \cite{carlucci2019domain,bucci2021self}. 
Meta-learning solutions get prepared for the source-target discrepancy experienced at test time by emulating the same condition with data drawn from the different sources during training \cite{li2018learning, li2019episodic}. 

In the \textit{Unsupervised Domain Adaptation} (UDA) setting the target data is available at training time but it is unlabeled. Possible strategies to close the domain gap are based on adversarial learning \cite{ganin2016domain} and feature alignment via MMD \cite{long2015learning} or via feature norms matching \cite{xu2019larger}. Pixel-wise adaptation can also be performed with
GAN-based techniques
\cite{CycleGAN2017, isola2017image}.
Clearly, MSDG and UDA share several solutions with slight differences  
due to the availability of multiple sources in one case, and source and target in the other. 
Finally, when the target is at least partially labeled, the setting is named \textit{Supervised Domain Adaptation} (SDA) 
and inherits most of the techniques developed for the more challenging UDA, SSDA and MSDA. Further constraints are eventually added to prevent overfitting in case of a very limited amount of labeled target data \cite{saito2019semi,li2021learning, yang2021deep, kim2020attract}.

As it is clear from the overview provided above, \emph{mitigating unfairness} and \emph{cross-domain learning} already share some solutions, which is possible by interpreting sensitive attributes as domains. The literature on fairness has been largely dedicated to evaluating tabular data with low-capacity models, while related works in computer vision focus only on  classification tasks. On the other hand, previous works on cross-domain learning broadly cover object classification and detection, as well as semantic segmentation, re-identification and retrieval problems \cite{csurka_book}. Still, the task of regression has been significantly less studied \cite{DAR_ICML_21,wu2022distributioninformed} and only a few works proposed robust methods for keypoint localization across domains \cite{2d_keypoints_ricci, jiang2021regda,keypoint_eccv2022,kim2022unified}. 

\vspace{1mm}\noindent\textbf{Landmark Detection}. 
Locating specific points in an image is crucial for applications like face recognition \cite{juhong_2017_facerecog, anghelone_2022_tfld}, object tracking \cite{huang_2022_stenttracking} and pose estimation \cite{xu2022vitpose}, with practical use in fields such as medicine, sports, and robotics. Keypoints as object corners and edges or facial features like the eyes, nose, and mouth are indicated as landmarks. 
Older landmark detection methods treat the task as a regression problem, where the goal is to predict 
continuous pixel coordinates 
for each landmark \cite{feng2018wing, wayne2018lab}. Recent methods have obtained significant gains in accuracy and robustness by modeling the landmark locations 
through a spatial probability distribution and providing high-resolution 2D heatmaps as output \cite{mccouat2022contour, jiang2021regda}. We consider these heatmap-based strategies in studying the problem of fair landmark detection.

\section{Fairness meets Cross-Domain Learning}
\begin{figure}[tb]
\centering
\includegraphics[width=0.93\linewidth]{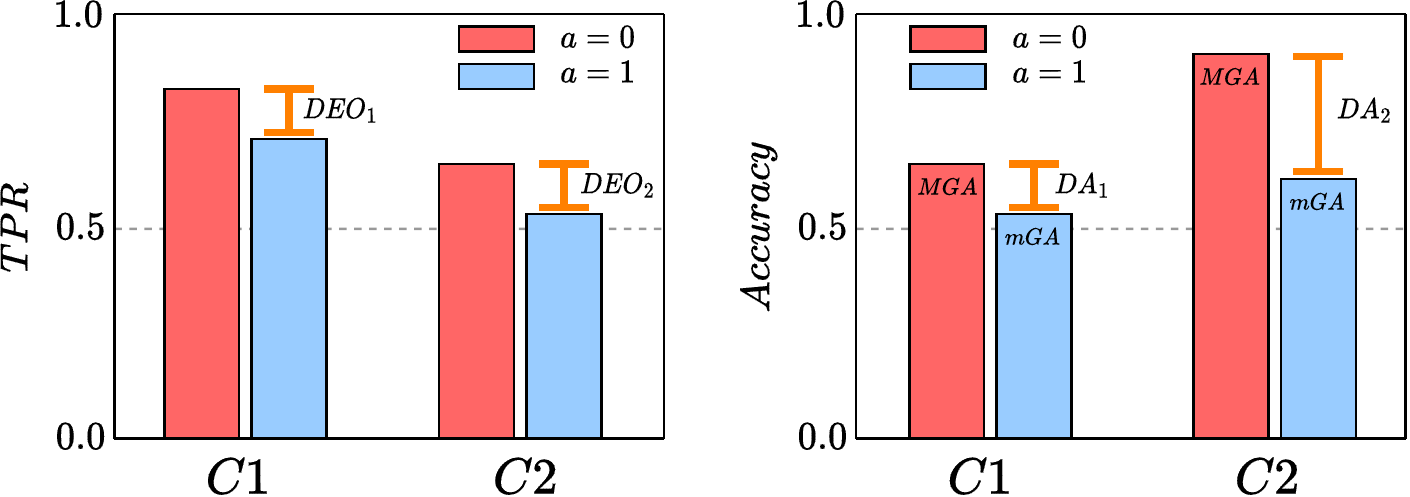}
\caption{Left: $C1$ and $C2$ have the same $DEO$ but $C1$ is clearly preferable to $C2$. Right: $C2$ has a higher $mGA$ than $C1$, but it also has a higher $DA$. Neither $DEO$ nor $mGA$ are sufficient for selecting a fair classifier with respect to sensitive attribute $a$. 
}\vspace{-4mm}
\label{fig:hist_fair}
\end{figure}

To formalize the problem of fairness we start by defining a data sample as $(\bx,y,a)$, where $\bx$ is an image, $y$ is its semantic label and $a$ is a sensitive attribute. 
In the simplest case, the labels are binary $y \in \{0,1\}$  (\eg for faces, eye bags  yes/no), and the same holds for the attributes $a \in \{0,1\}$ (\eg male/female or young/old). 
Given a set of annotated data spanning all the semantic labels and attributes, the goal is to learn a classifier $\hat{y}=f(\bx)$ that correctly predicts the label and achieves certain group fairness criteria with respect to $a$. 
These criteria mainly focus on the difference in performance between privileged and disadvantaged data groups associated with distinct attributes (see section \ref{sec:criteria}).

The presented fairness problem shares some common traits with that of cross-domain learning, where source $(\bx^s,y^s)\sim p^s$ and target $(\bx^t,y^t)\sim p^t$ data differ on the basis of the distribution from which they are drawn. The information about the distribution is usually summarized by a label indicating the data type: considering one source and one target domain, it holds  $d \in \{0,1\}$ (\eg photos/sketches). 
By simply switching $d$ with $a$ in the SDA setting we get to the framework described for the fairness problem.
As SDA can leverage the whole cross-domain literature, there is a large set of methods that can be applied and evaluated for unfairness mitigation. Some of them have been considered in previous fairness-related publications (\eg discrepancy, adversarial, and disentanglement strategies), but a thorough benchmark is still missing. As discussed in the following, letting cross-domain learning \emph{meet} fairness may lead to new evaluation strategies and interesting research questions.

\section{Fairness Criteria}\label{sec:criteria}

Evaluating the group fairness of a classification model means assessing its performance on different population subgroups and comparing them.
Many criteria have been proposed for this \cite{Verma2018fairness,Watcher2021bias}.
In the following, we review the most used metrics in computer vision. 
We start from the basic definitions of True Positive Rate $TPR=TP/(TP+FN)$, False Positive Rate
$FPR=FP/(FP+TN)$ and Accuracy $Acc = (TP+TN)/(TP+TN+FP+FN)$.
In terms of conditional probabilities for data 
with two different attributes, it holds 
\begin{align}
TPR_{a=0} & = P(\hat{y}=1|y=1, a=0) \\
FPR_{a=0} & = P(\hat{y}=1|y=0, a=0) 
\end{align}
and their analogues for $a=1$.
The \emph{Difference in Equal Opportunity (DEO)}  measures fairness by 
\begin{equation} 
|P(\hat{y}=1|y=1, a=0) - P(\hat{y}=1|y=1, a=1)|~,
\end{equation}
so the maximum fairness is obtained for $DEO=0$ when $TPR_{a=0}=TPR_{a=1}$. 
The \emph{Difference in Equalized Odds (DEOdds)}  measures fairness by 
\begin{equation} 
\sum_{t\in\{0,1\}}|P(\hat{y}=1|y=t, a=0) - P(\hat{y}=1|y=t, a=1)|~,
\end{equation}
thus maximum fairness is obtained for $DEOdds=0$ when both $DEO=0$ and $FPR_{a=0}=FPR_{a=1}$. In other words, the decision of the classifier should be conditionally independent of the attribute, given the ground truth ($\hat{y}\perp a | y$).
Another basic way to consider the variation of the model's output over the subgroups identified by the attributes is via the \emph{Difference in Accuracy (DA)}: 
\begin{equation}
|P(\hat{y}=y|a=0)-P(\hat{y}=y|a=1)|~.
\end{equation} 
All these metrics evaluate the relative behavior of the classifier on data subgroups defined by different attributes but lose track of its absolute performance. This is a critical issue as shown by the practical example in the left part of Figure \ref{fig:hist_fair}. Although the performance of the two classifiers is different, with $C1$ better than $C2$, they have the same value of $DEO$.
Moreover, both $DEO$ and $DEOdds$ are minimized by a trivial classifier that predicts always $\hat{y}=1$. In that case, for all the attributes it holds $FN=TN=0$, so $TPR=FPR=1$ and $DEO=DEOdds=0$. Since the accuracy reduces to the Positive Predictive Value ($PPV=TP/(TP+FP)$), also $DA$ becomes uninformative.

\begin{figure}[tb]
\centering
\includegraphics[width=0.93\linewidth]{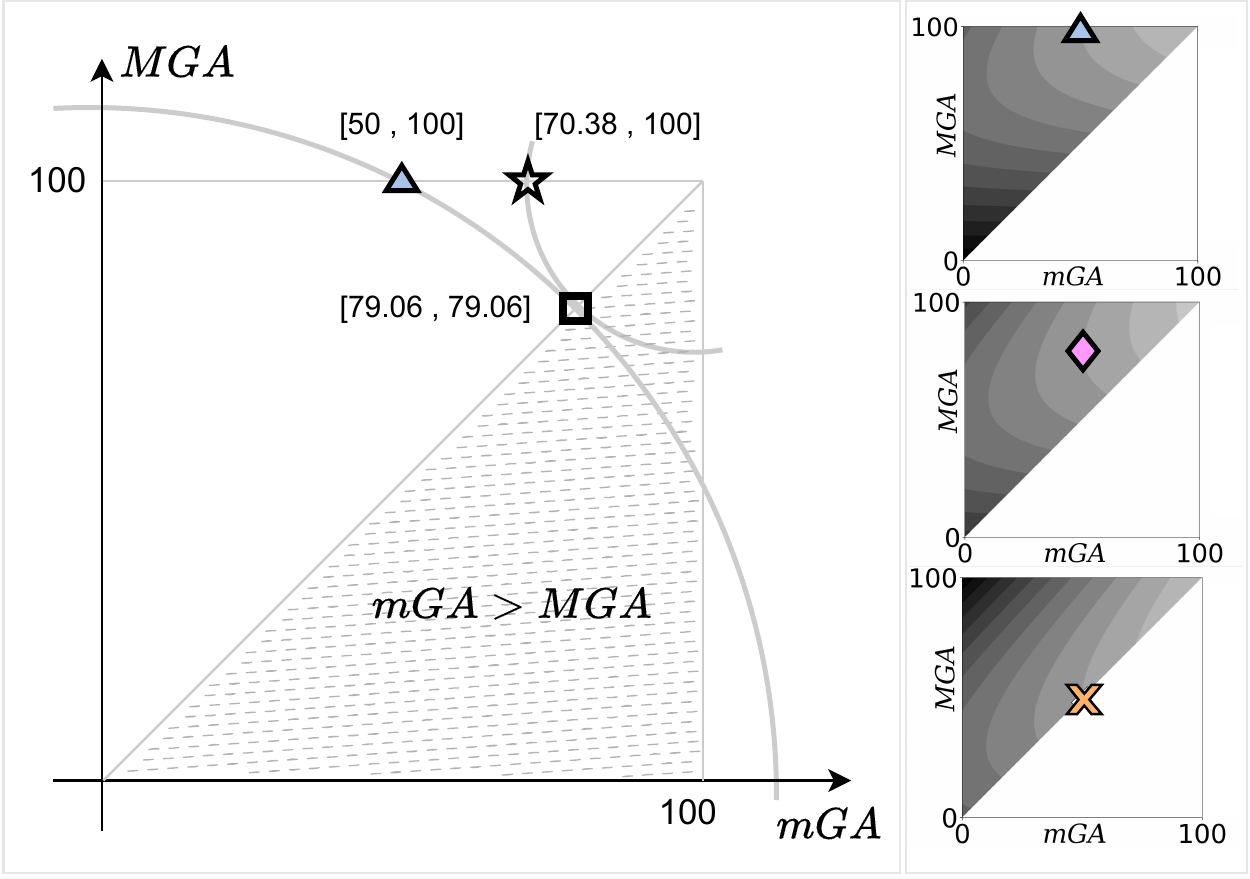}
\caption{Visualization of the $[mGA,MGA]$ space with exemplar points. The bottom triangular part of the space is unfeasible as by definition $mGA$ is lower than $MGA$. The three plots on the right show the HF isolines when starting from different baseline methods indicated by the $\triangle$, $\lozenge$ and $\times$ points. }
\label{fig:fairness_explained}
\vspace{-4.5mm}
\end{figure}

Recent works have introduced the \emph{Minimum Group Accuracy (mGA)} as fairness criterion: rather than evaluating differences in statistics across groups, it considers the classification accuracy of the worst performing group \cite{zietlow2022leveling,diana2021minimax,martinez2020minimax}. 
The rationale of this metric is that by increasing $mGA$ we are certainly improving the overall accuracy. Hence we avoid the suboptimal condition of unnecessarily harming all groups to get a trade-off improvement in fairness measured by $DEO$ and $DEOdds$.
Still, when the goal is to evaluate whether a certain unfairness mitigation method was able to improve over the reference classifier, $mGA$ is not sufficiently informative as exemplified by the right part of Figure \ref{fig:hist_fair}. Here $a=0$ is the privileged attribute, thus the one that identifies the best group with the associated \emph{Maximum Group Accuracy (MGA)}. When moving from $C1$ to $C2$, $mGA$ increases and so does $MGA$. Although globally the classifier has improved, the disadvantaged group suffers even more for unfair treatment with respect to the privileged one as indicated by the increased $DA$.

With these premises, we can state that fairness can be meaningfully assessed only together with prediction accuracy. Both their performance can be considered by looking at several bar plots jointly or at bi-dimensional plots as done in \cite{zietlow2022leveling}. However, interpreting them and making sense of multiple pieces of information at once is difficult, and defining a single score would facilitate rigorous quantitative evaluations. 
To this purpose, we can start from the space defined by $mGA$ and $MGA$. As shown in Figure \ref{fig:fairness_explained}, the bottom right triangular part of the space is an unfeasible region where  $mGA>MGA$. In the top right corner, the point with $[mGA,MGA]=[100,100]$ indicates the optimal utopia condition. The results of various methods can be collected in this space and ranked on the basis of the $L2$ Distance To the Optimum ($DTO$, \cite{zong2023medfair}) which sounds like a reasonable  metric for the score.
 
Let's focus for instance on the marked points in the figure and consider the biased reference classifier represented by $\triangle=[50,100]$. We expect a good unfairness mitigation method to keep the top $MGA=100$ result and improve $mGA$ to reduce the discrepancy among groups, thus moving horizontally towards the ideal point. The point $\medstar=[70.38,100]$ is a possible result for such an approach.
Differently, a method that trades off accuracy for fairness would decrease $MGA$ while improving $mGA$ to reduce $DA$ to zero. This behavior is exemplified by the 
point $\square=[79.06,79.06]$. It can be noticed that both $\square$ and $\triangle$ 
share the same Pareto efficiency level 
approximated by the circumference centered in $[0,0]$, as done in \cite{zietlow2022leveling}. Instead, $\medstar$ shows an advantage in efficiency, which is feasible as discussed in \cite{liu2023pushing}.
Despite their clear difference, the points $\medstar$ and $\square$ are equivalent according to $DTO$.
Thus, although $DTO$ keeps track of both $mGA$ and $MGA$ it might not be sufficiently informative to benchmark different unfairness mitigation approaches. The presented analysis also highlights the importance of taking as reference the performance of the
baseline to fully understand model comparisons.


\section{Harmonic Fairness}
\label{sec:HF2}

To better deal with the peculiarities of the space defined by $mGA$ and $MGA$, we \emph{formalize relative distances} for each method with respect to its biased reference and introduce \emph{our new Harmonic Fairness} metric. 

\vspace{1mm}\noindent\textbf{Classification.} 
We focus on $MGA$ and $DA=MGA-mGA$, using the subscripts $b$ and $m$ to refer respectively to the baseline model and its unfairness-mitigated version. The relative differences are: \vspace{-1mm}
\begin{align}\small
\Delta DA  & = DA_b - DA_m\\
\Delta MGA & = MGA_m - MGA_b
\end{align}
with $\Delta DA, \Delta MGA \in \{-100,100\}$. Both these values will be high for an accurate and fair model. Thus, we combine them in the \emph{Harmonic Fairness} metric defined as:
\begin{equation}\small
HF = \frac{\Delta DA' \times \Delta MGA'}{\Delta DA' + \Delta MGA'}~,
\label{eq:HF}
\end{equation}
where we added an additional shift to the component values to avoid degenerate cases (dividing by 0): $\Delta DA'=\Delta DA+100$ and $\Delta MGA'=\Delta MGA+100$.
The minimal value $HF=0$ corresponds to having either $\Delta DA = -100$ or $\Delta MGA = -100$, which can be obtained with a very poorly defined model that reduces the performance (increasing $DA$ or decreasing $MGA$) rather than improving over the baseline. An unfairness mitigation model that maintains the same $DA$ and $MGA$ of the original baseline gets $HF=50$. Finally, every increase over this value corresponds to models able to symmetrically improve accuracy and fairness. 
 
Getting back to the points $\medstar$ and $\square$ analyzed before and always considering the $\triangle$ as a baseline, we obtain the meaningful ranking $HF_{\medstar}=54.62 > HF_{\square}=51.77$ 
which matches the expectations given the advantage of the former over the latter. 
We remark that $HF$ takes into proper account the model starting baseline and encourages a decrease in $DA$ and an increase in $MGA$ with different strengths depending on the baseline position, consequently shaping the space in various ways as shown by the isolines of $HF$ in the right part of Figure \ref{fig:fairness_explained}. Of course, the right way to benchmark multiple methods is by setting a fixed baseline model considered as a shared reference for all of them.

\vspace{1mm}\noindent\textbf{Landmark Detection.}
When dealing with landmark detection every data sample can be defined as $(\bx,a,\bY)$, where $\bY\in \mathbb{R}^{K\times 2}$ is a set of $\by_{1,\ldots,K}$ landmark coordinates. The reference metric for this task is the Normalized Mean Error ($NME$) calculated as:\vspace{-1mm}
\begin{equation}\small
NME(\bY,\hat{\bY}) = \frac{1}{K}\sum_{i=1}^K\frac{\|\by_i-\hat{\by_i}\|_2}{D}~,
\end{equation}
where $D$ is a normalization factor, usually chosen as the interocular distance for face images. 
We indicate with $SDR$ the Success Detection Rate calculated as the percentage of images whose NMEs is less than a given threshold. Symmetrically to what was done for classification, we define \emph{Max Group Success} ($MGS$) and \emph{Min Group Success} ($mGS$), respectively as the success rate of the best and worst performing protected groups. We consider also the difference between groups $DS = MGS - mGS$, and to assess the effectiveness of an unfairness mitigation model $m$ over the reference baseline $b$ we calculate: \vspace{-1mm}
\begin{align}\small 
\Delta DS & = DS_b - DS_m \\
\Delta MGS & = MGS_m - MGS_b \vspace{-1mm}
\end{align}
with $\Delta DS, \Delta MGS \in \{-100,100\}$. We then combine these values to get the $HF$ metric for landmark detection consistent with what defined for classification in equation (\ref{eq:HF}).

\vspace{1mm}\noindent\textbf{Rescaling.}
To better investigates fine-grained differences among the results of various unfairness mitigation methods we adopt a simple sigmoid rescaling: $\sigma(HF) = \frac{1}{1+\exp^{-HF+50}}$, with $\sigma(HF) \in \{0,1\}$. Hence, $\sigma(HF)>0.5$ will indicate a gain over the reference baseline.

\section{Benchmark description}

For our analysis, we focus on two tasks that may be affected by fairness issues: attribute recognition and landmark detection. For the former, we consider two binary classification scenarios based on face and medical images.
For the latter, we focus on localizing keypoints on face images. To our knowledge, we are the first to study the impact of unfairness on landmark detection and to propose for it cross-domain learning as a possible solution.

\begin{table*}[t]
    \begin{center}
    \small
    \setcellgapes{1.5pt}
    \makegapedcells
    \resizebox{\textwidth}{!}{
    \begin{tabular}{|c|cccc|cc|c|c||cccc|cc|c|c|}
        \hline
        {} & \multicolumn{8}{c||}{\textbf{CelebA - EyeBags} (\emph{gender})} & \multicolumn{8}{c|}{\textbf{CelebA - Chubby} (\emph{gender})} \\
        \cline{2-17}
        {} & Acc. & \shortstack{MGA} & \shortstack{mGA} & \shortstack{DA} & \textcolor{gray}{DEO} & \textcolor{gray}{DEOdds}  & $\Delta DTO$ & $\mathbf{\sigma(HF)}$
        & Acc. & \shortstack{MGA} & \shortstack{mGA} & \shortstack{DA} & \textcolor{gray}{DEO} & \textcolor{gray}{DEOdds} & $\Delta DTO$ & $\mathbf{\sigma(HF)}$  \\ \hline
        
Baseline \cite{ramaswamy2021fair} & 81.46 & 88.59 & 70.15 & 18.44  & \textcolor{gray}{20.75} & \textcolor{gray}{39.10} & 0.00 & 0.500 $\pm$ 0.000
& 94.95 & 98.54 & 89.24 & 9.30 & \textcolor{gray}{27.55} & \textcolor{gray}{31.61} & 0.00 & 0.500 $\pm$ 0.000 \\ \hline

\textbf{LSR} \cite{szegedy2016rethinking} & 82.27 & 89.03 & 71.53 & 17.50 & \textcolor{gray}{22.67} & \textcolor{gray}{41.93} & 1.45 & 0.584	$\pm$ 0.034 
& 94.96 & 98.58 & 89.22 & 9.36 & \textcolor{gray}{27.37} & \textcolor{gray}{31.25} & -0.01  & 0.499 $\pm$ 0.014  \\

\textbf{SWAD} \cite{cha2021swad} & 80.25 & 87.24 & 69.15 & 18.09 & \textcolor{gray}{16.97} & \textcolor{gray}{44.67} & -1.42 & 0.444	$\pm$ 0.184 
& 94.20 & 98.47 & 87.43 & 11.04 & \textcolor{gray}{43.85} & \textcolor{gray}{51.67} & -1.80 & 0.393	$\pm$ 0.122 \\

\textbf{RSC} \cite{huang2020self} & 82.52 & 89.07 & 72.13 & 16.94 & \textcolor{gray}{15.44} & \textcolor{gray}{32.19}  & 2.02 & 0.621	$\pm$ 0.020 
& 94.89 & 98.40 & 89.33 & 9.06 & \textcolor{gray}{29.39} & \textcolor{gray}{33.98} & 0.06 & 0.506 $\pm$	0.013  \\

\textbf{L2D} \cite{lee2021learning} & 81.70 & 88.31 & 71.20 & 17.11 & \textcolor{gray}{18.31} & \textcolor{gray}{34.77} & 0.88 & 0.563 $\pm$	0.056 
& 95.07 & \underline{98.64} & 89.41 & 9.23 & \textcolor{gray}{28.30} & \textcolor{gray}{32.16}  & 0.18 & 0.510 $\pm$	0.006 \\
\hline

\textbf{DANN \cite{ganin2016domain}}  & \textbf{83.82} & \textbf{90.28} & \underline{73.56} & 16.72 & \textcolor{gray}{33.71} & \textcolor{gray}{48.89} & \underline{3.79} &  \underline{0.700} $\pm$	0.043 
& 95.07 & 98.56 & 89.53 & 9.03 & \textcolor{gray}{23.49} & \textcolor{gray}{26.81} & 0.29 & 0.518 $\pm$	0.011 \\

\textbf{CDANN \cite{li2018deep}} & 81.71 & 87.75 & 72.11 & 15.64 & \textcolor{gray}{\underline{8.11}} & \textcolor{gray}{\underline{22.91}} & 1.50 & 0.610	$\pm$ 0.078 
& 94.89 & 98.53 & 89.12 & 9.41  & \textcolor{gray}{36.05} & \textcolor{gray}{40.94} & -0.12 & 0.492	$\pm$ 0.020 \\

\textbf{SagNets} \cite{nam2021reducing} & 83.48 & \underline{90.17} & 72.85 & 17.32 & \textcolor{gray}{28.57} & \textcolor{gray}{45.82}   & 3.09 & 0.660	$\pm$ 0.075 
& \textbf{95.18} & \textbf{98.66} & \underline{89.65} & 9.01 & \textcolor{gray}{26.67} & \textcolor{gray}{30.35}  & \underline{0.42} & \underline{0.526} $\pm$	0.009\\
\hline

\textbf{AFN} \cite{xu2019larger} & \underline{83.59} & 89.34 & \textbf{74.47} & \underline{14.87} & \textcolor{gray}{\textbf{3.55}} & \textcolor{gray}{\textbf{12.05}} & \textbf{4.29} & \textbf{0.744} $\pm$	0.039 
& \underline{95.16} & 98.51 & \textbf{89.85} & \textbf{8.66} & \textcolor{gray}{22.58} & \textcolor{gray}{25.17}  & \textbf{0.60} & \textbf{0.536} $\pm$	0.021 \\

\textbf{MMD} \cite{li2018domain} & 83.53 & 89.94 & 73.35 & 16.59 & \textcolor{gray}{21.01} & \textcolor{gray}{36.57} 
& 3.47 & 0.689 $\pm$ 0.020 
& 95.11 & 98.55 & 89.64 & 8.90 & \textcolor{gray}{24.98} & \textcolor{gray}{28.69} & 0.40 & 0.525 $\pm$	0.030 \\

\textbf{Fish} \cite{shi2021gradient} & 82.45 & 89.43 & 71.38 & 18.05 & \textcolor{gray}{29.27} & \textcolor{gray}{51.81} & 1.45 & 0.575	$\pm$ 0.038 
& 94.95 & 98.57 & 89.19 & 9.39 & \textcolor{gray}{33.42} & \textcolor{gray}{37.84} & -0.04 & 0.496 $\pm$	0.030 \\
\hline

\textbf{RelRot} \cite{bucci2020effectiveness} & 82.88 & 89.55 & 72.29 & 17.26 & \textcolor{gray}{22.98} & \textcolor{gray}{42.66} & 2.35 & 0.629 $\pm$	0.063 
& 95.01 & 98.49 & 89.48 & 9.02 & \textcolor{gray}{26.99} & \textcolor{gray}{31.23} & 0.23 & 0.507 $\pm$	0.027 \\

\textbf{RelRotAlign} & 74.87 & 76.57 & 72.29 & \textbf{4.28} & \textcolor{gray}{33.69} & \textcolor{gray}{42.31} & -4.33 & 0.414	$\pm$ 0.113 
& 94.67 & 98.10 & 89.23 & \underline{8.87} & \textcolor{gray}{\textbf{2.04}} & \textcolor{gray}{\textbf{3.57}} & -0.08 & 0.493 $\pm$	0.007 \\

\textbf{SelfReg} \cite{kim2021selfreg} &  \underline{83.59} & 90.14 & 73.20 & 16.94 & \textcolor{gray}{21.78} & \textcolor{gray}{36.72} & 3.40 & 0.679 $\pm$	0.058 
& 95.07 & 98.61 & 89.46 & 9.15 & \textcolor{gray}{32.12} & \textcolor{gray}{36.09}  & 0.23  & 0.513 $\pm$	0.015 \\ 
\hline\hline

\textbf{GroupDRO} \cite{sagawa2019distributionally} & 83.08 & 89.25 & 73.28 & 15.98 & \textcolor{gray}{16.30} & \textcolor{gray}{30.82} & 3.16 & 0.681 $\pm$	0.067 
& 94.88 & 98.43 & 89.22 & 9.21 & \textcolor{gray}{30.14} & \textcolor{gray}{34.09} & -0.03 & 0.499 $\pm$	0.017  \\

\textbf{g-SMOTE} \cite{zietlow2022leveling} & 82.00 & 88.94 & 72.38 & 16.56 & \textcolor{gray}{28.11} & \textcolor{gray}{46.63} & 2.21 & 0.632	$\pm$ 0.071 
& 94.61 & 98.51 & 88.41 & 10.10 &\textcolor{gray}{27.25} & \textcolor{gray}{33.77} & -0.83 & 0.448	$\pm$ 0.026  \\

\textbf{FSCL} \cite{park2022fair} &  82.89 & 89.56 & 72.29 & 17.27 & \textcolor{gray}{34.51} & \textcolor{gray}{46.02} & 2.35 &  0.619 $\pm$	0.192 
& 95.00 & 98.48 & 89.48 & 9.00 & \textcolor{gray}{\underline{15.53}} & \textcolor{gray}{\underline{19.42}} & 0.23 & 0.518	$\pm$ 0.010  \\ \hline
    \end{tabular}
    }
    \end{center}
    \vspace{-2mm}
    \caption{Results obtained on face images when the task is to recognize \textit{EyeBags} (left) and \textit{Chubby} (right) with gender as sensitive attribute. Every number represents the average over three runs. \textbf{Bold} indicates the best results, \underline{underline} the second best. }
    \label{tab:celebaA_EB_C} \vspace{-4mm}
\end{table*}

\subsection{Datasets}
\noindent\textbf{CelebFaces Attribute (CelebA)} \cite{liu2015faceattributes}
comprises 202,599 RGB face images of celebrities, each with 40 binary attribute annotations. 
We focus on the same subset of 13 reliable target attributes considered in \cite{ramaswamy2021fair,zietlow2022leveling}.
We select \textit{male} and \textit{young} as protected attributes, and adopt the same setting of \cite{zietlow2022leveling}, based on the official train/val/test splits. 

\noindent\textbf{COVID-19 Chest X-Ray} \cite{cohen2020covidProspective} is composed of 719 images of chest x-ray coming from different online sources showing scans of patients affected by pulmonary diseases. Each image has a structured label describing many attributes of the patient.  
We focus on the \textit{finding} attribute as target, considering the COVID-19 pathology, while \textit{gender} is selected as sensitive attribute. 
We split the dataset into 80/20\% training/test sets, using 20\% of the training split for validation.

\noindent\textbf{Fitzpatrick17k} \cite{groh2021evaluating} is a collection of 16,577 clinical images depicting 114 skin conditions from two dermatology atlases.
The images are annotated with the six Fitzpatrick skin type labels, that describe the skin phenotype's sun reactivity. The dataset is widely used in algorithmic fairness research \cite{zong2023medfair}. 
We classify whether the dermatological condition in each picture is either \emph{benign/non-neoplastic} or \emph{malignant} and we use \emph{skin tone} as the protected attribute, keeping only the examples belonging to \emph{skin type I} (light) and \emph{skin type VI} (dark) of the Fitzpatrick scale. We split the dataset into 80/20\% training/test sets, using 20\% of the training split for validation.

\noindent\textbf{UTK Face} \cite{zhang2017age} consists of over 20k RGB face images
characterized by great variability in terms of pose, facial expression, illumination, \etc., and present age, gender, and race annotations.
We focus on landmark localization (68 points) considering the values \emph{white} and \emph{black} of the label \textit{race}
as protected groups for the experiments related to \emph{skin tone}. Moreover, we define the \emph{young} and \emph{old} groups by collecting respectively samples with the value of label \emph{age} in 0-10 and 40-50 years old.
Training/test division is in the proportion 80/20\% with 20\% of the training split used for validation.

\subsection{Reference Methods}
\noindent\textbf{Baselines.} 
For our classification experiments we follow the fairness literature \cite{zietlow2022leveling,sagawa2019distributionally} adopting as baseline ResNet50 with standard cross-entropy minimization objective, pre-trained on ImageNet. 
For landmark detection we follow \cite{jiang2021regda,simple_baseline_landmark} and consider ResNet18 pre-trained on ImageNet with a dedicated head composed of deconvolutional layers. It is optimized with an $L2$ loss to reduce the discrepancy between the predicted probability distribution of the location of each landmark and the ground truth.

\noindent\textbf{Fairness References.}
We consider three SOTA  
unfairness mitigation methods. GroupDRO~\cite{sagawa2019distributionally}
minimizes the worst-case training loss over a set of pre-defined groups. FSCL~\cite{park2022fair} re-designs supervised contrastive learning to ensure fairness by paying attention to the choice of the negative samples and to the distribution of the anchors between data groups. Finally, g-SMOTE~\cite{zietlow2022leveling} is a generative approach that reduces unfairness by synthesizing new samples of the most disadvantaged group.
All of them focus on classification problems while we are not aware of works dedicated to unfairness mitigation on landmark detection. 

\noindent\textbf{Cross-Domain Models.}
We investigate methods from four main families. 
The \emph{regularization-based approaches} include all the techniques designed to prevent overfitting with a consequent boost in the model generalization ability. 
LSR~\cite{szegedy2016rethinking} encourages the model to avoid overconfidence by smoothing data annotation.
SWAD~\cite{cha2021swad} searches for flat minima. 
RSC~\cite{huang2020self} is based on a refined drop-out.
L2D~\cite{wang2021learning} includes a module trained to synthesize new images with a style distribution complementary to that of the training data.
The models based on \emph{Adversarial training} encode domain-invariant representations by preventing the network to recognize the domains.
In DANN~\cite{ganin2016domain} the gradient computed by a domain discriminator is inverted while learning the data representation. 
CDANN~\cite{li2018deep} improves over DANN by matching the conditional data distributions across domains rather than the marginal distributions. 
Finally, SagNets~\cite{nam2021reducing} introduces dedicated data randomizations to disentangle style from class encodings.
\emph{Feature alignment} models involve training objectives that minimize domain distance measures.
AFN~\cite{xu2019larger} measures domain shift by comparing the feature norms of two domains and adapts them to a common large value. 
MMD~\cite{sejdinovic2013equivalence} minimizes the homonym metric to reduce the domain discrepancy. 
Lastly, Fish~\cite{shi2021gradient} proposes to align the domain distributions by maximizing the inner product between their gradients.
\emph{Self-Supervised Learning}-based techniques exploit auxiliary self-supervised tasks to let the network focus on semantic-relevant features.
RelRot~\cite{bucci2020effectiveness} predicts the relative orientation between a reference image (anchor) and the rotated counterpart as auxiliary task.
Here we also consider a variant
that we name RelRotAlign to encourage the domain alignment using as anchor a sample with the same target attribute but from a different protected group. SelfReg~\cite{kim2021selfreg}, exploited contrastive losses to regularize the model and guide it to learn domain-invariant representations.

\begin{table*}[t]
    \begin{center}
    \small
    \setcellgapes{1.5pt}
    \makegapedcells
    \resizebox{\textwidth}{!}{

    \begin{tabular}{|c|cccc|cc|c|c||cccc|cc|c|c|}
        \hline
        {} & \multicolumn{8}{c||}{\textbf{COVID-19 Chest X-Ray} (\emph{gender})}
        & \multicolumn{8}{c|}{\textbf{Fitzpatrick17k} (\emph{skin tone})}\\
        \cline{2-17}
        {} & Acc. & \shortstack{MGA} & \shortstack{mGA} & \shortstack{DA} & \textcolor{gray}{DEO} & \textcolor{gray}{DEOdds}  & $\Delta DTO$ & $\mathbf{\sigma(HF)}$      
        & Acc. & \shortstack{MGA} & \shortstack{mGA} & \shortstack{DA} & \textcolor{gray}{DEO} & \textcolor{gray}{DEOdds} & $\Delta DTO$  & $\mathbf{\sigma(HF)}$ \\ \hline
     
Baseline   \cite{ramaswamy2021fair}    & 73.79 & 78.62 & 65.41 & 13.22 & \textcolor{gray}{26.44} & \textcolor{gray}{35.18} & 0.00  & 0.500 $\pm$ 0.000 
& 90.22 & 94.45 & 87.26 & 7.19 & \textcolor{gray}{11.51} & \textcolor{gray}{13.29} & 0.00 & 0.500 $\pm$ 0.000  
\\ \hline

\textbf{LSR} \cite{szegedy2016rethinking} &  68.96 & 72.83 & 62.26 & 10.57 & \textcolor{gray}{22.41} & \textcolor{gray}{26.83} & -5.84   & 0.307 $\pm$	0.144 
 &  \underline{92.89} & 93.66 & \underline{92.25} & \underline{1.41} & \textcolor{gray}{19.65} & \textcolor{gray}{20.53} & \underline{3.88} & \underline{0.766} $\pm$ 0.020    
\\

\textbf{SWAD} \cite{cha2021swad} &  73.10 & \underline{79.35} & 62.26 & 17.09 & \textcolor{gray}{19.49} & \textcolor{gray}{25.48} & -2.36   & 0.348 $\pm$	0.278 
&  91.59 & 93.20 & 90.44 & 2.76 & \textcolor{gray}{10.31} & \textcolor{gray}{11.29} & 2.16 & 0.678 $\pm$	0.054 
\\

\textbf{RSC} \cite{huang2020self} &  \textbf{77.93} & \textbf{80.43} & \textbf{73.58} & 6.85 & \textcolor{gray}{\underline{8.62}} & \textcolor{gray}{\underline{11.32}} & \textbf{7.79}  & \textbf{0.880} $\pm$	0.035   
&  91.72 & 92.70 & 90.89 & 1.81 & \textcolor{gray}{20.63} & \textcolor{gray}{21.07}  & 2.22 & 0.698 $\pm$ 0.046
\\

\textbf{L2D} \cite{lee2021learning} & 70.34 & 71.74 & 67.92 & 3.82 & \textcolor{gray}{12.07} & \textcolor{gray}{21.14} & -2.09   & 0.574 $\pm$	0.049  
& 92.56 & \textbf{94.77} & 90.91 & 3.86 & \textcolor{gray}{16.52} & \textcolor{gray}{17.59} & 3.41 & 0.709 $\pm$ 0.021
\\

\hline
\textbf{DANN} \cite{ganin2016domain}  & 71.03 & 71.74 & 69.81 & \textbf{1.93} & \textcolor{gray}{\underline{8.62}} & \textcolor{gray}{\textbf{10.83}} & -0.69 & 0.666 $\pm$	0.068   
& 92.04 & 92.98 & 91.32 & 1.66 & \textcolor{gray}{11.38} & \textcolor{gray}{13.44} & 2.73 & 0.720 $\pm$ 0.046
\\

\textbf{CDANN} \cite{li2018deep} & 70.34 & 72.83 & 66.04 & 6.79 & \textcolor{gray}{17.24} & \textcolor{gray}{22.88} & -2.83  & 0.492	$\pm$ 0.116 
& 91.85 & 92.95 & 90.99 & 1.95 & \textcolor{gray}{\underline{7.40}} & \textcolor{gray}{\underline{9.15}} & 2.46 & 0.704	$\pm$ 0.045
\\

\textbf{SagNets} \cite{nam2021reducing} & 74.48 & 79.35 & 66.04 & 13.31  & \textcolor{gray}{15.52} & \textcolor{gray}{27.04}  & 0.92 & 0.540 $\pm$ 	0.042 
& 91.46 & 92.70 & 90.58 & 2.12 & \textcolor{gray}{15.10} & \textcolor{gray}{15.86} & 1.98 & 0.683 $\pm$ 0.033
\\
\hline

\textbf{AFN} \cite{xu2019larger} &  \underline{76.55} & \underline{79.35} & \underline{71.70} & 7.65 & \textcolor{gray}{13.79} & \textcolor{gray}{16.98}  & 5.63   & 0.821 $\pm$	0.045 
  &  91.95 & 93.60 & 90.72 & 2.87 & \textcolor{gray}{15.42} & \textcolor{gray}{16.74} & 2.62 & 0.694 $\pm$ 0.033 
\\

\textbf{MMD} \cite{li2018domain} & \underline{76.55} & \textbf{80.43} & 69.81 & 10.62 & \textcolor{gray}{13.79} & \textcolor{gray}{24.58} & 4.69 & 0.741 $\pm$	0.113   
& 91.60 & 93.30 & 90.43 & 2.87 & \textcolor{gray}{8.84} & \textcolor{gray}{10.28} & 2.21  & 0.679 $\pm$ 0.018
\\

\textbf{Fish} \cite{shi2021gradient} & 75.86 & 78.26 & \underline{71.70} & 6.56 & \textcolor{gray}{17.24} & \textcolor{gray}{24.10}  & 4.98 & 0.812 $\pm$		0.072   
& 92.24 & 93.52 & 91.39 & 2.12 & \textcolor{gray}{17.14} & \textcolor{gray}{18.79} & 3.12  & 0.723 $\pm$ 0.078
\\
\hline

\textbf{RelRot} \cite{bucci2020effectiveness} & 74.48 & 76.09 & \underline{71.70} & 4.39 & \textcolor{gray}{10.34} & \textcolor{gray}{16.47} & 3.62  & 0.793 $\pm$	0.093  
& \textbf{93.35} & 94.47 & \textbf{92.61} & 1.87 & \textbf{\textcolor{gray}{2.01}} & \textbf{\textcolor{gray}{2.81}} & \textbf{4.67} & \textbf{0.786} $\pm$ 0.017  
\\

\textbf{RelRotAlign} & 75.86 & 77.17 & \textbf{73.58} & \underline{3.59} &\textcolor{gray}{12.07} & \textcolor{gray}{22.36}  & \underline{5.75} &   \underline{0.857} $\pm$	0.073 
& 91.65 & 92.81 & 90.70 & 2.11 & \textcolor{gray}{12.02} & \textcolor{gray}{13.36} & 2.14  &  0.687 $\pm$ 0.039
\\

\textbf{SelfReg} \cite{kim2021selfreg} & 73.79 & 77.17 & 67.92 & 9.25 & \textcolor{gray}{17.24} & \textcolor{gray}{18.71} & 1.29 & 0.642 $\pm$	0.067   
& 92.69 & \underline{94.61} & 91.27 & 3.34 & \textcolor{gray}{8.34} & \textcolor{gray}{10.14} & 3.64 & 0.727 $\pm$ 0.029
\\ 
\hline\hline

\textbf{GroupDRO} \cite{sagawa2019distributionally} & 70.34 & 73.91 & 64.15 & 9.76 &\textcolor{gray}{22.41} & \textcolor{gray}{29.77} & -3.67 & 0.403 $\pm$	0.095  
& 91.98 & 92.64 & 91.41 & \textbf{1.24} & \textcolor{gray}{10.19} & \textcolor{gray}{11.60} & 2.58  & 0.722 $\pm$ 0.037
\\

\textbf{g-SMOTE} \cite{zietlow2022leveling} & 73.14 & 77.97 & 64.11 & 13.86  & \textcolor{gray}{25.60} & \textcolor{gray}{34.49} & -1.45  & 0.420 $\pm$ 0.043 
 & 90.92 & 93.19  &  88.50 & 4.69 & \textcolor{gray}{12.26} & \textcolor{gray}{14.27} & 0.53  & 0.569 $\pm$ 0.066
\\

\textbf{FSCL} \cite{park2022fair} & 60.02 & 63.04 & 54.86 & 8.18  & \textcolor{gray}{\textbf{1.72}} & \textcolor{gray}{15.20} & -17.68 &  0.051 $\pm$	0.035  
 & 90.67 & 93.60 & 88.48 & 5.13 & \textcolor{gray}{12.42} & \textcolor{gray}{13.85} & 0.72 & 0.566 $\pm$ 0.176
\\ \hline
        
    \end{tabular}
    }
    \end{center}
    \vspace{-2mm}
    \caption{Results on medical images for covid recognition, with gender as sensitive attribute. Avrage results over three runs.
    }
    \label{tab:covid+melanoma}
    \vspace{-4mm}

\end{table*}

\noindent\textbf{Landmark Detection.}
The community has dedicated less attention to domain adaptive approaches for keypoint detection. For our analysis, we consider the recent RegDA~\cite{jiang2021regda} that was developed to target human pose estimation and introduced an adversarial regressor based on the 
Kullback-Leibler divergence between domains to narrow their gap.
We also extend  DANN~\cite{ganin2016domain} and AFN~\cite{xu2019larger} to this task.

\section{Experiments}

In this section we present the main results of our experiments. Further evaluations are added in the supplementary material which also includes all the implementation details as well as the code. Our PyTorch implementation covers all the methods evaluated in the benchmark to guarantee maximal transparency and reproducibility. It can also easily include other methods for future benchmark extensions. Unless stated otherwise, for all the experiments 
we adopted the same validation protocol described in \cite{zietlow2022leveling}.

\subsection{Classification Results}
For the binary classification tasks, we organize the tables with different horizontal sections that group the cross-domain methods by family. 
The bottom part of the tables contains the SOTA fairness approaches. 
Besides the standard metrics and our newly introduced $\sigma(HF)$ we consider
$\Delta DTO = DTO_b - DTO_m$: as the baseline is fixed and shared by all the methods, $\Delta DTO$ ranks the methods exactly as $DTO$ but makes the tables easier to read.

\textbf{The results on CelebA} are presented in Table \ref{tab:celebaA_EB_C} and focus on the two most challenging attributes: \emph{EyeBags} and \emph{Chubby}. Out of the whole set of 13 attributes (complete results in the supplementary), they are the ones with the highest $DA$ and lowest $Acc$. 
By focusing on both $\sigma(HF)$ and $\Delta DTO$ we can state that several cross-domain methods provide an accuracy and fairness gain over the baseline, and are also able to improve over the SOTA fairness methods which appear particularly inefficient on Chubby. The AFN approach shows the best performance, followed by DANN on Eyebags and SagNets on Chubby.  

Regarding the metrics,
$\Delta DTO$ and $\sigma(HF)$ 
agree on the general ranking. On the other hand, neither $mGA$ nor $DA$ is sufficiently informative. For instance on Eyebags, RelRot and RelRotAlign have the same $mGA$, while both their $\Delta DTO$ and $\sigma(HF)$ are significantly different, with RelRotAlign even worsening the baseline. Moreover, the best $DA$ result of RelRotAlign is clearly misleading as it comes with a noteworthy decrease in Acc. 
Finally, we notice 
how $DEO$ and $DEOdds$ for Chubby reward the \emph{leveling down} behavior already criticized in \cite{zietlow2022leveling}, by largely relying on $DA$ without considering the decrease in $MGA$ and $mGA$ with respect to the baseline.  

\textbf{The results on COVID-19 Chest X-Ray and Fizpatrick17} are presented in Table \ref{tab:covid+melanoma}. 
On the first dataset, 
according to both $\Delta DTO$ and $\sigma(HF)$, RSC is the top method and RelRotAlign is the second best, while AFN ranks third. 
Interestingly, now 
the rankings of DANN and SagNets differ depending on the used metric: 
$\sigma(HF)$ rewards the former, more than the latter. DANN has a significant advantage in $mGA$ despite a loss in $MGA$,  while SagNets shows a minimal increase in both $mGA$ and $MGA$ in spite of a worse discrepancy among the groups.
Even in this case it is clear that referring only to $mGA$ may not be sufficient to differentiate among the methods as many of them share the exact same value for this metric. 

The results on Fizpatrick17 lead to similar conclusions, with RelRot and LSR presenting the best results. DANN, which was among the top methods for CelebA, now ranks sixth among all the CD approaches and still shows results comparable with the best SOTA fairness approach.

Overall, the exact the family that best suits each classification task may vary (feature alignment and adversarial training methods for faces, regularization-based and self-supervised approaches for medical images), but the results confirm the effectiveness of cross-domain learning for unfairness mitigation and the relevance of our study.

\begin{figure}
\begin{tabular}{cc} 
\hspace{-7.5mm}
\includegraphics[width=0.55\linewidth]{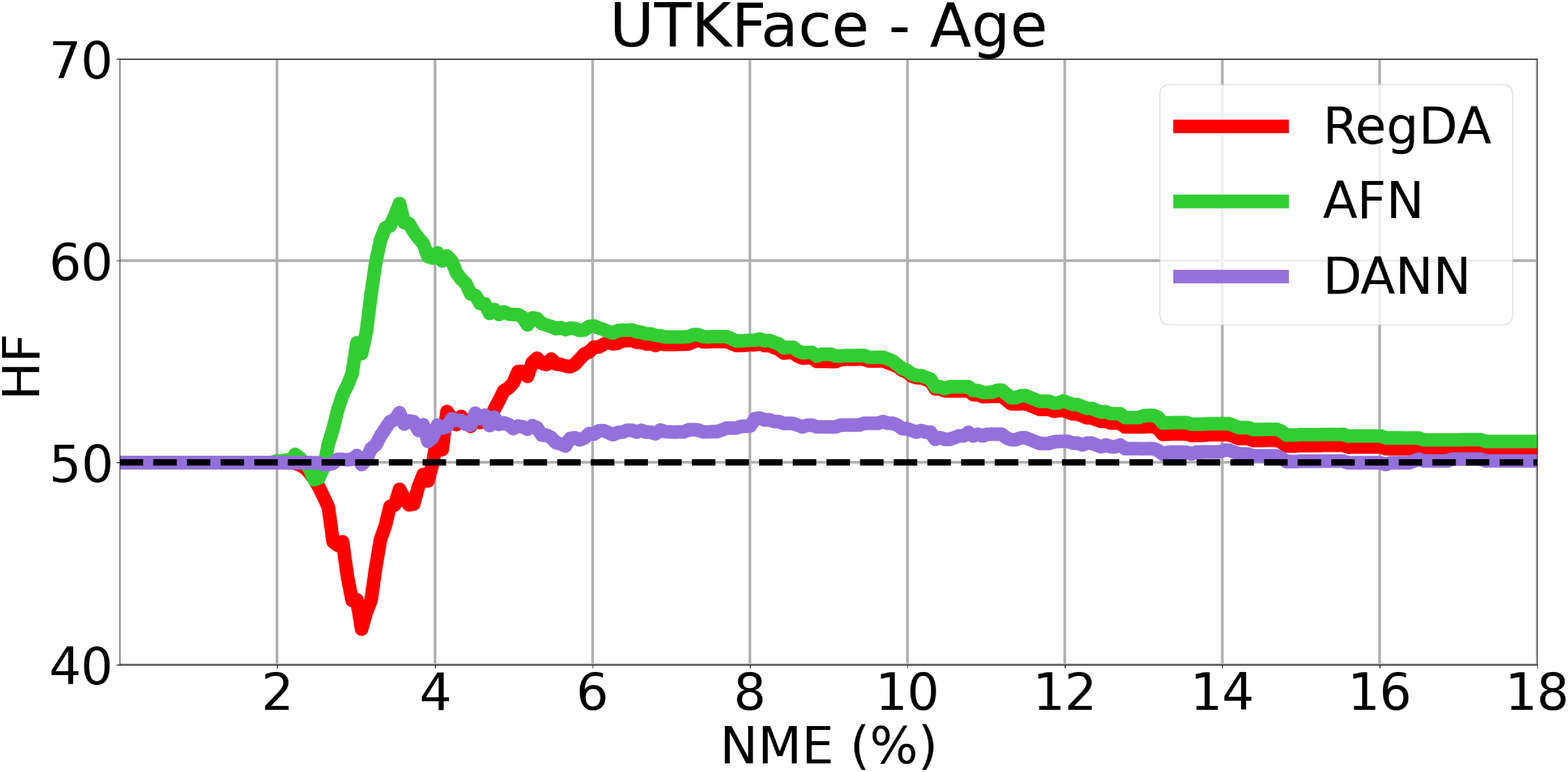} &
\hspace{-0.5cm}
\includegraphics[width=0.55\linewidth]{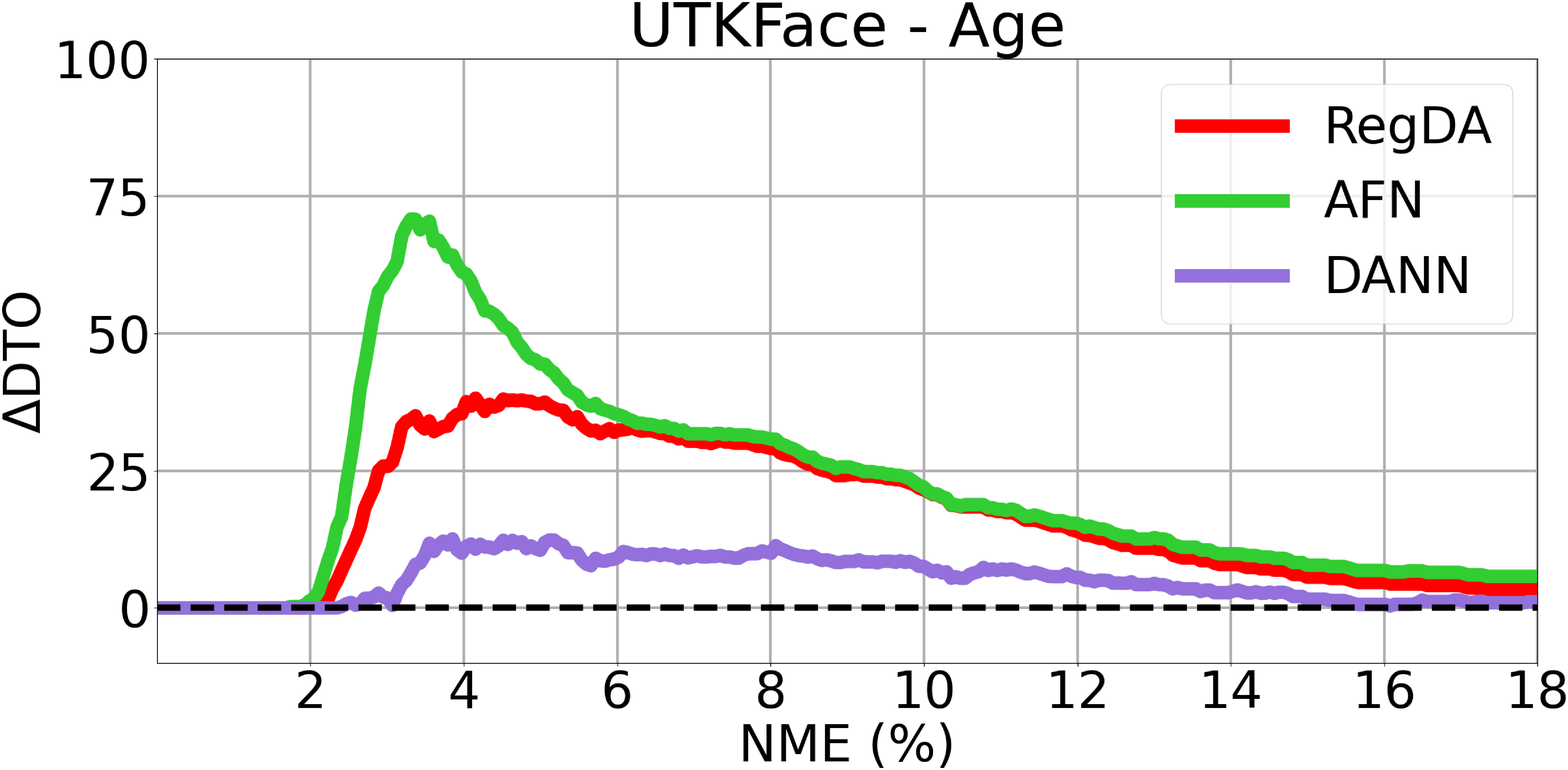}
\\
\end{tabular}
\caption{Landmark detection results. Comparison among the cross-domain methods and the reference baseline in terms of $HF$ and $\Delta DTO$ when changing the NME threshold used for $SDR$.}
\label{fig:NMEth} \vspace{-2mm}
\end{figure}

\begin{table}[t]
    \begin{center}
    \small
    \resizebox{0.48\textwidth}{!}{
    \begin{NiceTabular}{|c | c c c c | c | c | }[colortbl-like,cell-space-limits=1.5pt]
        \hline
        {} & \multicolumn{6}{c|}{\textbf{UTK Face Landmark} (\emph{skin tone})} 
        \\
        \cline{2-7}
        {} & SDR & \shortstack{MGS} & \shortstack{mGS} & \shortstack{DS} &  $\Delta DTO$ & $\mathbf{\sigma(HF)}$ 
        \\ \hline
        
Baseline  & 82.08 & 83.90 & 77.93 & 5.97 & 0.00 & 0.500 $\pm$ 0.000  
\\ \hline

\textbf{AFN} \cite{xu2019larger} & \underline{92.95} & \underline{93.80} & \underline{90.99} & \textbf{2.81} & \underline{16.38} & \underline{0.961} $\pm$ 0.047 
\\

\textbf{DANN \cite{ganin2016domain}} & 89.62 & 90.66 & 86.17 & 4.49 & 10.63 & 0.883 $\pm$ 0.017 
\\ 

\textbf{RegDA}  \cite{jiang2021regda} & \textbf{96.05} & \textbf{97.05} & \textbf{93.77} & \underline{3.28} & \textbf{20.43} & \textbf{0.979} $\pm$ 0.011  
\\ 

\hline

{} & \multicolumn{6}{c|}{\textbf{UTK Face Landmark} (\emph{age})} \\
\hline
Baseline  & 74.50 & 78.39 & 70.71 & 7.67 & 0.00 & 0.500 $\pm$ 0.000 \\
\hline
\textbf{AFN}  \cite{xu2019larger}& \underline{94.50} & \underline{95.04} & \textbf{93.97} & \textbf{1.06} & \underline{28.59} & \textbf{0.997} $\pm$ 0.001 \\
\textbf{DANN \cite{ganin2016domain}} & 81.03 & 85.22 & 76.83 & 8.39 & 8.92 & 0.809 $\pm$ 0.091 \\
\textbf{RegDA}  \cite{jiang2021regda} & \textbf{94.62} & \textbf{95.51} & \underline{93.74} & \underline{1.77} & \textbf{28.70} & \underline{0.996} $\pm$ 0.001  \\
\hline

    \end{NiceTabular}
    }
    \end{center}
    \vspace{-2mm}
    \caption{Landmark detection results. SDR is evaluated using 8\% NME as threshold. Results averaged over three runs.
    }
    \vspace{-4mm}
    \label{tab:landmark}
\end{table}

\subsection{Landmark Detection Results}
The performance of a model which locates keypoints on facial components may be affected by a change in \emph{skin tone} and \emph{age}, resulting in a less precise prediction in case of high melanin pigmentation or wrinkles. To investigate the presence of a bias related to these demographics we run experiments on the UTK Face dataset and we verify the effectiveness of correction strategies based on cross-domain learning by considering RegDA together with AFN and DANN, as they have shown successful results in classification on face images. The training procedure follows the one presented in \cite{jiang2021regda}, with validation protocol in line with that of \cite{zietlow2022leveling}. 
We assess the performance of the methods by considering both our $\sigma(HF)$ and $\Delta DTO$ obtained from $SDR$ calculated with a standard 8\% $NME$ threshold \cite{mccouat2022contour}. 

Table \ref{tab:landmark} shows how the baseline reference has an unfair behavior with more than 5\% difference in group accuracy ($DS$). All the cross-domain methods provide an advantage: in particular, RegDA ranks higher or equal to AFN, and they are both better than DANN. The latter shows a large improvement in $MGS$ and $mGS$ when the sensitive attribute is age, but the group discrepancy appears worse than the baseline.  
By reducing the $NME$ threshold the evaluation becomes progressively more demanding until the extreme of considering a predicted point as successful only if it perfectly overlaps with the ground truth. The curves in Figure \ref{fig:NMEth} show that even moving toward this condition most of the cross-domain methods maintain their advantage over the baseline confirming their effectiveness.  
The difference between RegDA and AFN becomes more evident at lower threshold values. In that regime $HF$ (as well as $\sigma(HF)$) and $\Delta DTO$ show different trends for RegDA with the first discouraging the use of this approach.

\begin{table}[t]
    \begin{center}
    \small
    \resizebox{0.48\textwidth}{!}{
     \begin{NiceTabular}{|c | c c c c | c | c |}[colortbl-like,cell-space-limits=1.5pt]
        \hline
        {} & \multicolumn{6}{c}{\textbf{CelebA - EyeBags}}\\
        \hline
        {} & \multicolumn{6}{c|}{{\textit{Male/Female} $\longrightarrow$ \textit{Young/Old}}}
        \\
        \cline{2-7}
        {} & Acc. & \shortstack{MGA} & \shortstack{mGA} & \shortstack{DA} & $\Delta DTO$ & $\mathbf{\sigma(HF)}$
        \\ \hline

Baseline \cite{wang2020towards} & 81.06	& 83.20 & 74.05 & 9.16 & 0.00 & 0.500 $\pm$ 0.000 
\\ \hline

\textbf{AFN} \cite{xu2019larger} & 81.31	& 83.30 & 74.92 & \textbf{8.38} & 0.78 & 0.554 $\pm$	0.014  
\\ 

\textbf{DANN} \cite{ganin2016domain}  & \textbf{83.48} &	\textbf{85.83} &	\textbf{76.16} &	9.67 & \textbf{3.18} & \textbf{0.626} $\pm$	0.039
\\ 

\hline
\textbf{GroupDRO} \cite{sagawa2019distributionally} & \underline{82.52} &	\underline{84.90} &	\underline{75.68} &	\underline{9.22} & \underline{2.29}  &   \underline{0.599} $\pm$	\underline{0.055}
\\ 

\textbf{g-SMOTE} \cite{zietlow2022leveling}  & 80.16 &	82.58 &	72.62 &	9.96  & -1.54 & 0.415 $\pm$	0.168 
\\ 

\textbf{FSCL} \cite{park2022fair}  & 80.45 &	84.65 &	69.35 &	15.30	& -3.37 &   0.235 $\pm$	0.097  
\\ 

\hline

\rowcolor[gray]{0.95}
         & \multicolumn{6}{c|}{\textit{Young/Old} $\longrightarrow$ \textit{Young/Old} (\textbf{Oracle})}
         \\
        \hline

\rowcolor[gray]{0.95}
\textbf{AFN} \cite{xu2019larger} & 81.70	& 83.81 & 75.04 & 8.77 & 1.16 & 0.561 $\pm$	0.013 
\\ 

\rowcolor[gray]{0.95}
\textbf{DANN} \cite{ganin2016domain}  & 83.00 &	84.90 &	77.01 &	7.89 & 3.41 & 0.676 $\pm$ 	0.040 
\\ 
\hline

\rowcolor[gray]{0.95}
\textbf{GroupDRO}  \cite{sagawa2019distributionally}  & 82.23 & 83.70 & 75.67 & 8.03 & 1.63 & 0.598 $\pm$	0.049 
\\ 

\rowcolor[gray]{0.95}
\textbf{g-SMOTE} \cite{zietlow2022leveling}  & 	81.21 &	81.99 &	73.71 &	8.28 & -0.95 & 0.477 $\pm$	0.072  
\\ 

\rowcolor[gray]{0.95}
\textbf{FSCL} \cite{park2022fair}  & 80.50 &	84.66 &	69.39 &	15.27	& -3.33 &  0.237 $\pm$		0.098 
\\ 

 \hline
        
    \end{NiceTabular}
    }
    \end{center}
    \vspace{-2mm}
    \caption{Model Transferability analysis on the classification task. All the relative metrics are calculated with respect to the baseline results in the first row.}
    \label{tab:celeba_transfer}
    \vspace{-4mm}
\end{table}

\section{Model Transferability}
Considering the effort needed to train novel models, it is always desirable to exploit existing ones for new tasks.  
For unfairness mitigation approaches, what is learned by reducing the bias over some protected groups might be helpful also for other demographics. We study this aspect on the CelebA dataset considering \emph{EyeBags} as the target attribute with \emph{Male} and \emph{Young} as sensitive attributes.
We train and validate a classifier to recognize whether eye bags are present while learning to disregard gender-specific features through a cross-domain approach. Then, we test the obtained model by assessing how the eye bags prediction performance differs among age groups. We analyze the CD methods AFN and DANN, reporting also the results of the SOTA unfairness mitigation strategies. 
The top part of Table \ref{tab:celeba_transfer} shows the effect on age groups of the approaches trained to be gender agnostic while focusing on the semantic features relevant to identify the target \emph{EyeBags} attribute. The results exceed those of the baseline with a particular advantage of DANN over GroupDRO, indicating that the knowledge acquired with cross-domain learning is easily transferrable.
The bottom part of the table presents the performance of \emph{oracle} methods trained and validated with the aim of mitigating age bias. They represent an upper bound and allow to better appreciate the surprisingly competitive results of transferred cross-domain models.
We note also how the SOTA unfairness mitigation models obtain low results even in this oracle setting.
More experiments with inverted roles for the sensitive attributes (\textit{Young/Old} $\rightarrow$ \textit{Male/Female}) and similar settings on the landmark detection task are reported and discussed in the supplementary.

\section{Conclusions}
In this paper we proposed an extensive study on the problem of fairness in computer vision by presenting a new benchmark 
to assess the performance of cross-domain learning approaches for unfairness mitigation. Moreover, we introduced the task of landmark detection in the fairness research area and  proposed the Harmonic Fairness. This new metric takes into account both the accuracy and the degree of fairness of a model to evaluate its effectiveness. Although our focus is mainly on group fairness and other definitions are possible, we believe that our work provides several tools to broaden the study of fairness-related issues and solutions in AI. 

\smallskip\noindent\textbf{Acknowledgments.} Computational resources were provided by IIT (HPC infrastructure) and CINECA through the IsC98\_FA-DA project under the ISCRA initiative. L.I. acknowledges the grant received from the European Union Next-GenerationEU (Piano Nazionale di Ripresa E Resilienza (PNRR)) DM 351 on Trustworthy AI. S. B. acknowledges the travel grant provided by the Blanceflor Foundation.  This work has been also partially supported (T.T.) by the EU project ELSA - European Lighthouse on Secure and Safe AI (\url{https://www.elsa-ai.eu/}). 

{\small
\bibliographystyle{ieee_fullname}
\bibliography{egbib}
}


%

\newpage 
\appendix


\setcounter{section}{0}
\setcounter{table}{0}
\setcounter{figure}{0}




\section{Implementation Details}

\subsection{Classification}
For all the experiments we follow \cite{zietlow2022leveling} in terms of base architecture, training details, and validation protocol. In particular, all the methods are built upon the ImageNet pre-trained ResNet-50 \cite{he2016deep} backbone optimized with Adam ($lr=10^{-4}$, batch size 64). 
As data augmentation, we use a center crop to 128x128 and RandAugment with $N=3$ and $M=15$. The validation is done every 500 iterations and the best model is selected based on the best $mGA$ computed on the validation set. Note that for g-SMOTE \cite{zietlow2022leveling} we used the GAN inversion model provided in  \cite{dinh2021hyperinverter}, pre-trained on CelebA: the official GAN code and weights used in \cite{zietlow2022leveling} have not been released by the authors.
Although there may be some debate around the use of generative approaches that are not tailored specifically to the medical task at hand, we decided to incorporate g-SMOTE into both the experiments on the COVID-19 Chest X-Ray and Fitzpatrick17k datasets for completeness.

We perform an extensive hyper-parameters search to find the best models for every approach considered in our benchmark. In particular, we apply the Random Search \cite{bergstra2012random} algorithm followed by a refinement stage in the following hyper-parameter intervals:

\smallskip\noindent \textbf{LSR} \cite{szegedy2016rethinking}: $\varepsilon$ is the coefficient used to smooth the ground truth labels such that $y_k^{LS}=y_k(1-\varepsilon)+\varepsilon/K$, where $K$ indicates the number of classes. $ \left \{ \varepsilon \in [0.1, 0.5] \right \} $ ;

\smallskip\noindent  \textbf{SWAD} \cite{cha2021swad}: $r$ is the tolerance rate used on the validation loss function when searching the interval in which the model's parameters have to be sampled and averaged. We didn't tune the optimum patience ($N_e$) and the overfit patience ($N_s$) since the overfitting behavior could be observed already after the very first validation. $ \left \{ r \in [0.1, 1.3] \right \}$ ;
 
\smallskip\noindent \textbf{RSC} \cite{huang2020self}: $f$ indicates the dropping percentage to mute the spatial feature maps, $b$ indicates the percentage of the batch on which RSC is applied. $ \left \{  f \in [10, 80] ,  b \in [10, 80] \right \}$ ;

\smallskip\noindent\textbf{L2D} \cite{lee2021learning}: $\alpha_1$ weights the contribution of the supervised contrastive loss function and $\alpha_2 $ weights the negative log-likelihood between the latent vectors of the source image $x$ and the generated one $x^+$ in the final objective function. $ \left \{  \alpha_1 \in [0.1, 3.0], \alpha_2 \in [0.1, 3.0] \right \} $ ;

\smallskip\noindent \textbf{DANN} \cite{ganin2016domain}, \textbf{CDANN} \cite{li2018deep}: $\lambda$ is the hyper-parameter that weights the reverse gradient during the backpropagation step, $\gamma$ controls the penalty assigned to the norm computed on the gradients of the domain discriminator.  $ \left \{ \lambda \in [0.01, 1.00], \gamma \in [0.01, 0.50] \right \} $ ;
 
\smallskip\noindent\textbf{SagNets} \cite{nam2021reducing}: $\lambda$ weights the adversarial loss function. $  \left \{ \lambda \in [0, 2] \right \} $ ;

\smallskip\noindent \textbf{AFN} \cite{xu2019larger}:  $\lambda$ trades off the feature-norm penalty and the supervised cross-entropy loss, $R$ is the value at which the norms of the extracted features are forced to converge to. $ \left \{ \lambda \in [0.01, 0.10], R \in [5, 100] \right \}$ ;
 
\smallskip\noindent \textbf{MMD} \cite{li2018domain}: $\gamma$ weights the MMD loss term in the final objective. $ \left \{ \gamma \in [0.1, 1.0] \right \} $ ;
 
\smallskip\noindent \textbf{Fish} \cite{shi2021gradient}: $\eta$ weights the gradient inner product.  $ \left \{  \eta \in [0.01, 0.10] \right \}$ ;

\smallskip\noindent \textbf{RelRot},  \textbf{RelRotAlign} \cite{bucci2020effectiveness}: $\alpha$ weights the importance of the self-supervised loss function in the total objective. $ \left \{  \alpha \in [0.1, 1.0] \right \} $ ;

\smallskip\noindent \textbf{SelfReg} \cite{kim2021selfreg}: $\lambda_{feature}$ and $\lambda_{logit}$ control, respectively, the in-batch dissimilarity losses applied to the intermediate features and the logits from the classifier. $ \left \{  \lambda_{feature} \in [0.1, 1.0], \lambda_{logit} \in [0.1, 1.0] \right \} $ ;
 
\smallskip\noindent \textbf{GroupDRO} \cite{sagawa2019distributionally}: $C$ is a model capacity, $\eta$ is the step size to update the weights in order to balance worst and best performing groups. $ \left \{  \eta \in [0.001, 0.05],  C \in [1, 10] \right \} $ ;

\smallskip\noindent \textbf{g-SMOTE} \cite{zietlow2022leveling}: $m$ in the number of nearest neighbors considered, $k$ is the number of random points chosen among the $m$ and $\lambda$ is the probability of selecting a batch from the original dataset during training. $ \left \{ m \in [2, 10], k \in [2, m], \lambda \in [0.1, 1.0] \right \}$ .


\subsection{Landmark Detection}

Throughout our experiments, we adopt the architecture and training procedures outlined in \cite{jiang2021regda}. To ensure consistency, we also use the validation protocol proposed in \cite{zietlow2022leveling}. Our approach employs an ImageNet pre-trained ResNet-18 \cite{he2016deep} backbone, followed by an upsampling head consisting of three 2D transposed convolutions with a dimension of 200 and a kernel size of 4. This head performs heatmap regression to determine the position of each landmark, resulting in an output tensor $\mathbf{\hat{Y}} \in \mathbb{R}^{200 \times 200 \times 68}$.
Our network is optimized using stochastic gradient descent (SGD) with a learning rate of 0.1, momentum of 0.9, weight decay of 1e-4, and a batch size of 32 for 35000 iterations. We incorporate a multi-step learning rate decay with a decay factor of 0.1, using iteration 22500 and 30000 as milestones.
To apply several augmentation sequentially we use the TorchLM library\footnote{\url{https://github.com/DefTruth/torchlm}}.
The augmentations are: random rotation (with angles ranging from -180 to 180 degrees), random horizontal flip (with a probability of 0.5), random shear (with x and y rescale factors of 0.6 and 1.3, respectively), color jitter (with brightness, contrast, and saturation set to 0.24, 0.25, and 0.25, respectively) and Gaussian blur (with a kernel size of 5 and $\sigma=(0.1, 0.8)$). We validate every 500 iterations and select the best model based on the highest $mGS$ score on the validation set.

We conduct an exhaustive search for optimal hyperparameters for all the approaches included in our benchmark. Specifically, we employ the Random Search algorithm \cite{bergstra2012random}, followed by a refinement stage, within the hyperparameter intervals as specified below:

\smallskip\noindent \textbf{AFN} \cite{xu2019larger}:  $\lambda$ trades off the feature-norm penalty and the supervised cross-entropy loss, $R$ is the value at which the norms of the extracted features are forced to converge to. $ \left \{ \lambda \in [1e-6, 0.10], R \in [5, 100] \right \}$ ;

\smallskip\noindent \textbf{DANN} \cite{ganin2016domain}: $\lambda$ is the hyper-parameter that weights the reverse gradient during the backpropagation step, $\gamma$ controls the penalty assigned to the norm computed on the gradients of the domain discriminator.  $ \left \{ \lambda \in [1e-6, 1.00], \gamma \in [0.01, 0.50] \right \} $;

\smallskip\noindent \textbf{RegDA} \cite{jiang2021regda}: \emph{margin} trades off between the KL divergence loss and the Regression Disparity loss. $t$ is a modifier of the magnitude of the Regression Disparity loss. $ \left \{ margin \in [1.0, 10.0], t \in [0.01, 1.0] \right \} $.

\section{CelebA - 13 Attributes Experiment}
\begin{table}[t]
    \begin{center}
    \small
    \setcellgapes{1.5pt}
    \makegapedcells
    \resizebox{0.48\textwidth}{!}{
    \begin{tabular}{|c | c c c c | c c | c | c |}
        \hline
        {} & \multicolumn{8}{c|}{\textbf{CelebA - 13 Attributes} (\emph{gender})} 
        \\
        \cline{2-9}
        {} & Acc. & \shortstack{MGA} & \shortstack{mGA} & \shortstack{DA} & \textcolor{gray}{DEO} & \textcolor{gray}{DEOdds} & $\Delta DTO$ & $\mathbf{\sigma(HF)}$  \\ \hline
        
Baseline \cite{ramaswamy2021fair} & 91.90 & 93.71 & 89.60 & 4.12 & \textcolor{gray}{15.05} & \textcolor{gray}{18.47} & 0.00 & 0.500 $\pm$ 0.000 \\ \hline

\textbf{LSR} \cite{szegedy2016rethinking} & 91.97 & 93.76 & 89.72 & 4.04 & \textcolor{gray}{14.85} & \textcolor{gray}{18.27} & 0.13 & 0.507 $\pm$ 0.010  \\

\textbf{SWAD} \cite{cha2021swad} & 91.45 & 93.50 & 88.92 & 4.58 & \textcolor{gray}{17.23} & \textcolor{gray}{22.20} & -0.69 & 0.458 $\pm$ 0.034  \\

\textbf{RSC} \cite{huang2020self} & 91.89 & 93.57 & 89.80 & \underline{3.77} & \textcolor{gray}{14.44} & \textcolor{gray}{17.81} & 0.10 & 0.513 $\pm$	0.007  \\

\textbf{L2D} \cite{lee2021learning} & 91.91 & 93.74 & 89.66 & 4.08 & \textcolor{gray}{14.91} & \textcolor{gray}{18.03} & 0.07 & 0.504 $\pm$	0.008  \\
\hline

\textbf{DANN \cite{ganin2016domain}}  & \textbf{92.15} & \underline{93.92} & \underline{89.96} & 3.95 & \textcolor{gray}{16.02} & \textcolor{gray}{19.08} & \underline{0.42} & \underline{0.523} $\pm$ 	0.001\\

\textbf{CDANN \cite{li2018deep}} & 92.04 & 93.73 & 89.92 & 3.80 & \textcolor{gray}{13.46} & \textcolor{gray}{16.25} & 0.28 & 0.520 $\pm$	0.016   \\

\textbf{SagNets} \cite{nam2021reducing} & 92.08 & 93.87 & 89.86 & 4.00 & \textcolor{gray}{17.14} & \textcolor{gray}{20.58} & 0.31 &  0.516	$\pm$ 0.014 \\
\hline

\textbf{AFN} \cite{xu2019larger} & \underline{92.14} & 93.85 & \textbf{90.02} & 3.83 & \textcolor{gray}{13.24} & \textcolor{gray}{15.97} & \textbf{0.43} & \textbf{0.527} $\pm$	0.008  \\

\textbf{MMD} \cite{li2018domain} & 92.09 & 93.83 & 89.93 & 3.90 & \textcolor{gray}{15.24} & \textcolor{gray}{18.35} & 0.34 & 0.521 $\pm$	0.008  \\

\textbf{Fish} \cite{shi2021gradient} & 91.85 & 93.69 & 89.58 & 4.11 & \textcolor{gray}{15.55} & \textcolor{gray}{19.02} & -0.03 & 0.499 $\pm$	0.000  \\
\hline

\textbf{RelRot} \cite{bucci2020effectiveness} & 92.10 & 93.86 & 89.90 & 3.95 & \textcolor{gray}{15.47} & \textcolor{gray}{18.97} & 0.33 & 0.519 $\pm$	0.006    \\

\textbf{RelRotAlign} & 91.14 & 92.39 & 89.70 & \textbf{2.69} & \textcolor{gray}{\textbf{8.29}} & \textcolor{gray}{\textbf{10.50}} & -0.65 & 0.503 $\pm$	0.014   \\

\textbf{SelfReg} \cite{kim2021selfreg} & 91.98 & 93.76 & 89.75 & 4.01 & \textcolor{gray}{15.50} & \textcolor{gray}{18.18} & 0.15 & 0.510 $\pm$	0.011  \\ 
\hline\hline

\textbf{GroupDRO} \cite{sagawa2019distributionally} & 91.97 & 93.78 & 89.77 & 4.01 & \textcolor{gray}{14.05} & \textcolor{gray}{17.10} & 0.18 & 0.511	$\pm$ 0.005   \\

\textbf{g-SMOTE} \cite{zietlow2022leveling} & 92.12 & \textbf{93.94} & 89.88 & 4.05 & \textcolor{gray}{14.68} & \textcolor{gray}{18.47} & 0.36 & 0.517	$\pm$ 0.007  \\

\textbf{FSCL} \cite{park2022fair} & 91.58 & 93.50 & 89.14 & 4.36 & \textcolor{gray}{\underline{13.22}} & \textcolor{gray}{\underline{15.78}} & -0.50 & 0.473 $\pm$ 0.002  \\ 
\hline
        
    \end{tabular}
    }
    \end{center}
\caption{Results obtained on CelebA considering the whole set of 13 reliable attributes as target, while gender is the sensitive attribute.  The numbers represent the average over 13 experiments, each repeated 3 times. \textbf{Bold} indicates the best results, \underline{underline} the second best.}
    \label{tab:celeba13_5}
\end{table}

\begin{figure}
\begin{tabular}{cc} 
\hspace{-7.5mm}
\includegraphics[width=0.55\linewidth]{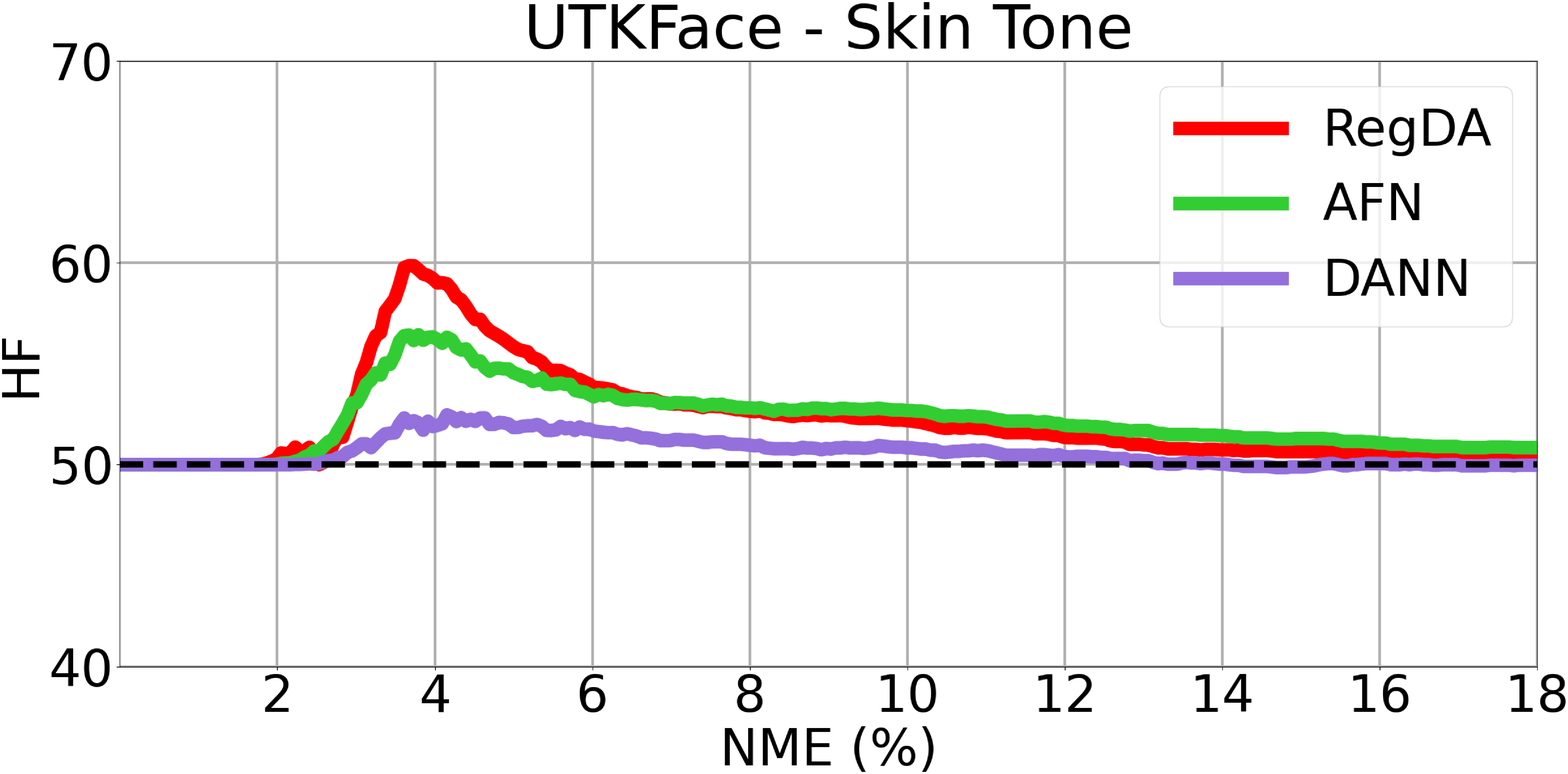} & 
\hspace{-0.5cm}
\includegraphics[width=0.55\linewidth]{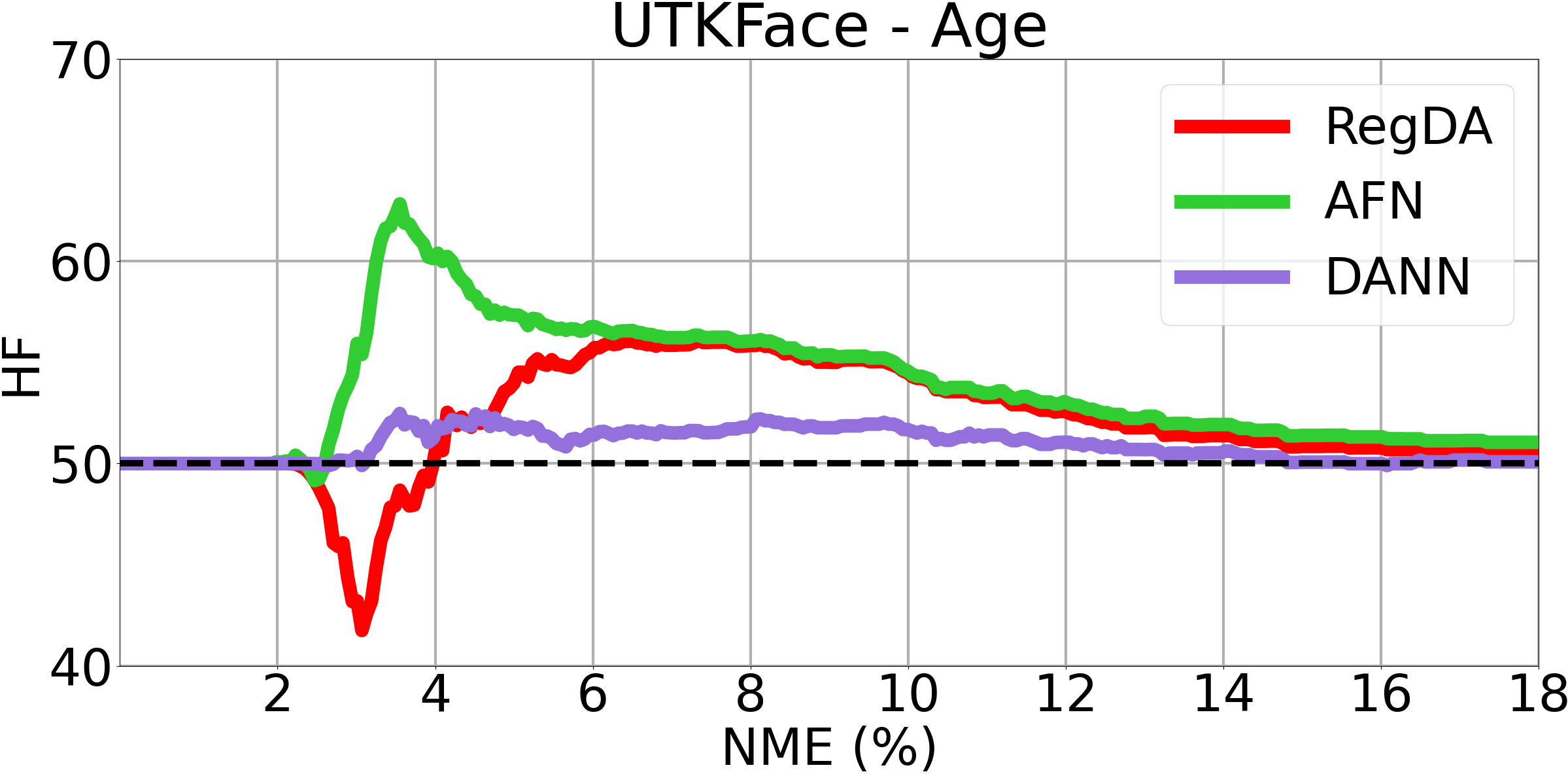}
\\
\hspace{-7.5mm}
\includegraphics[width=0.55\linewidth]{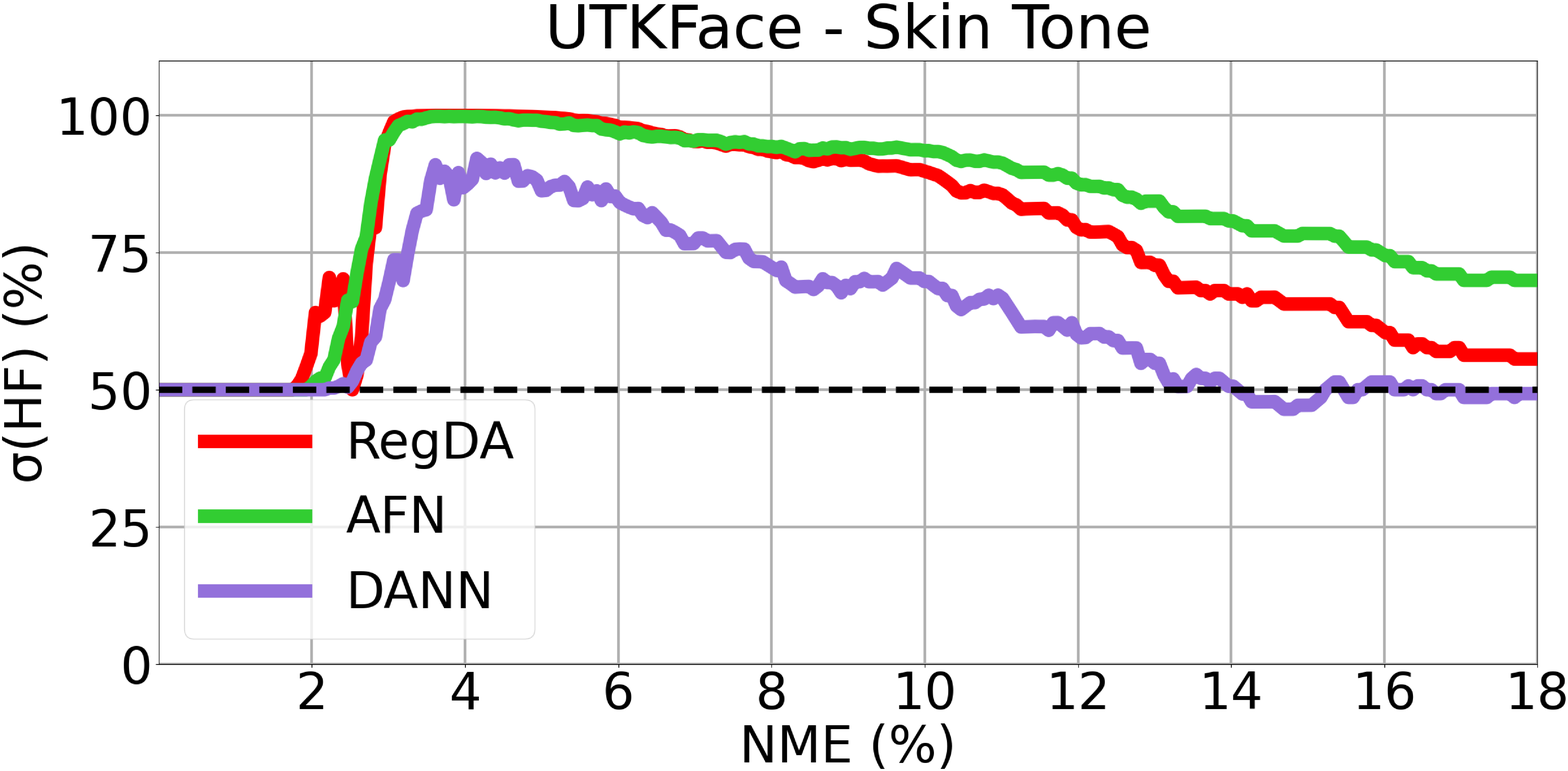} &  \hspace{-0.5cm}
\includegraphics[width=0.55\linewidth]{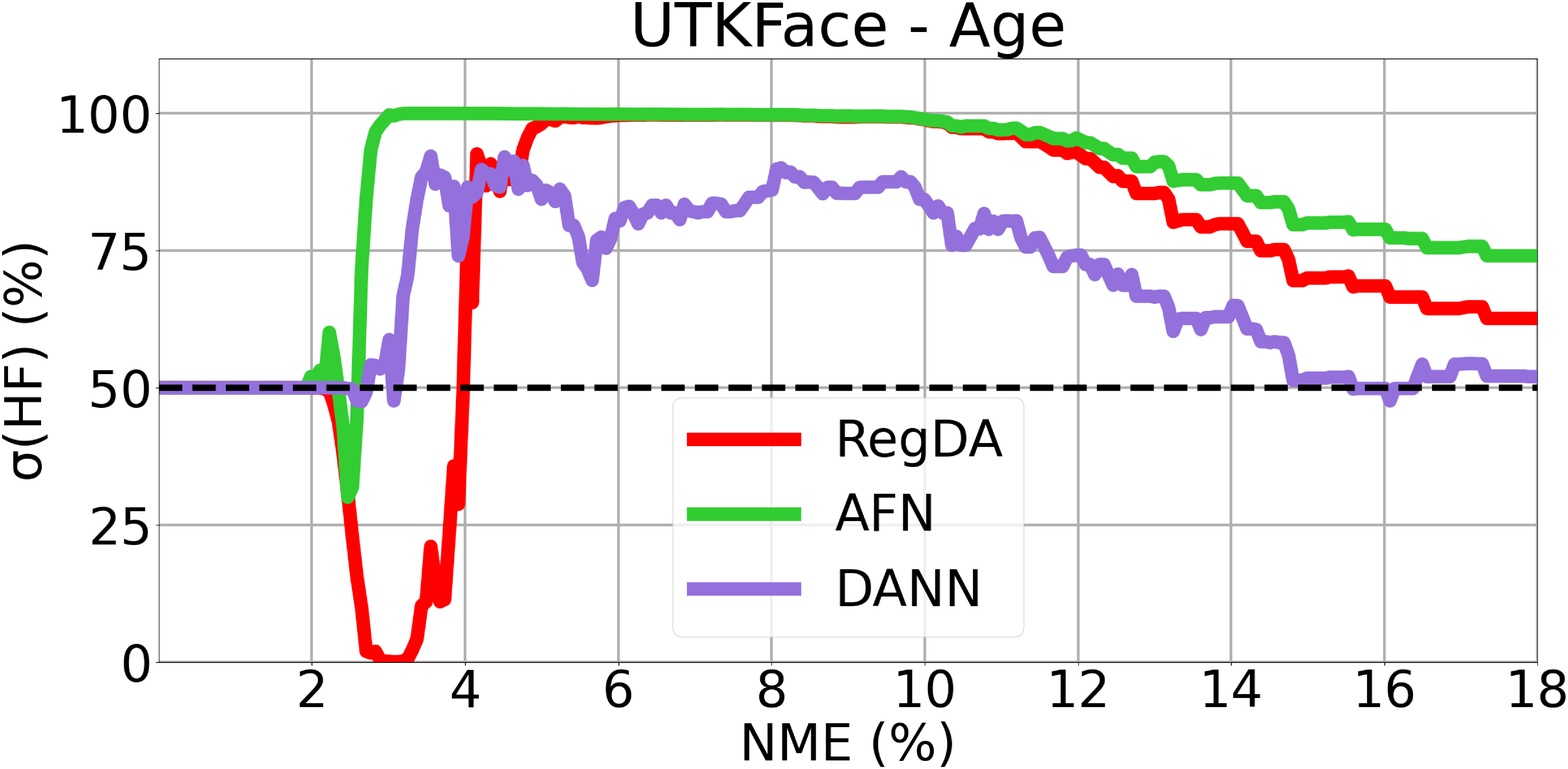}
\\
\hspace{-7.5mm}
\includegraphics[width=0.55\linewidth]{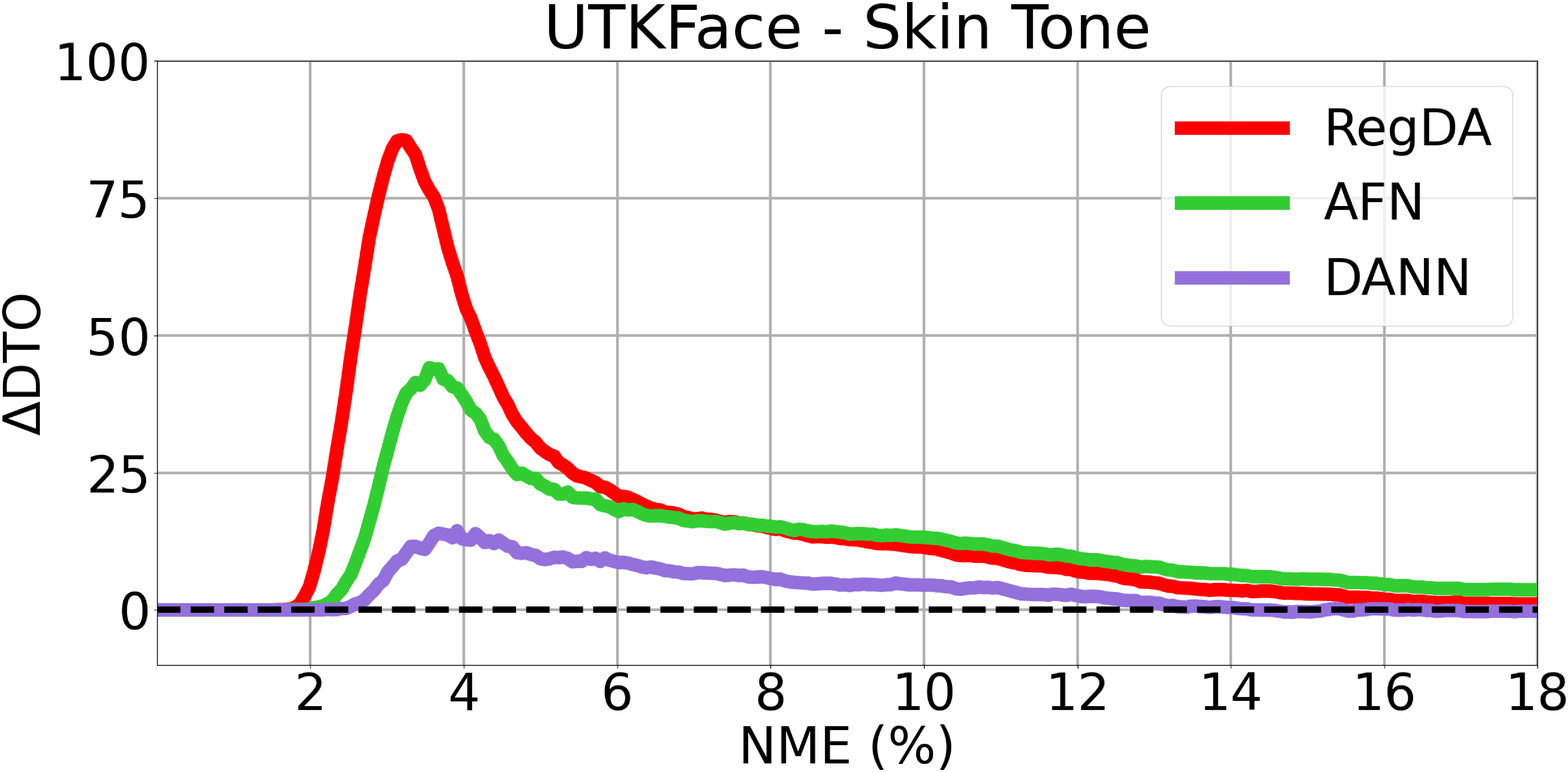} &  \hspace{-0.5cm}
\includegraphics[width=0.55\linewidth]{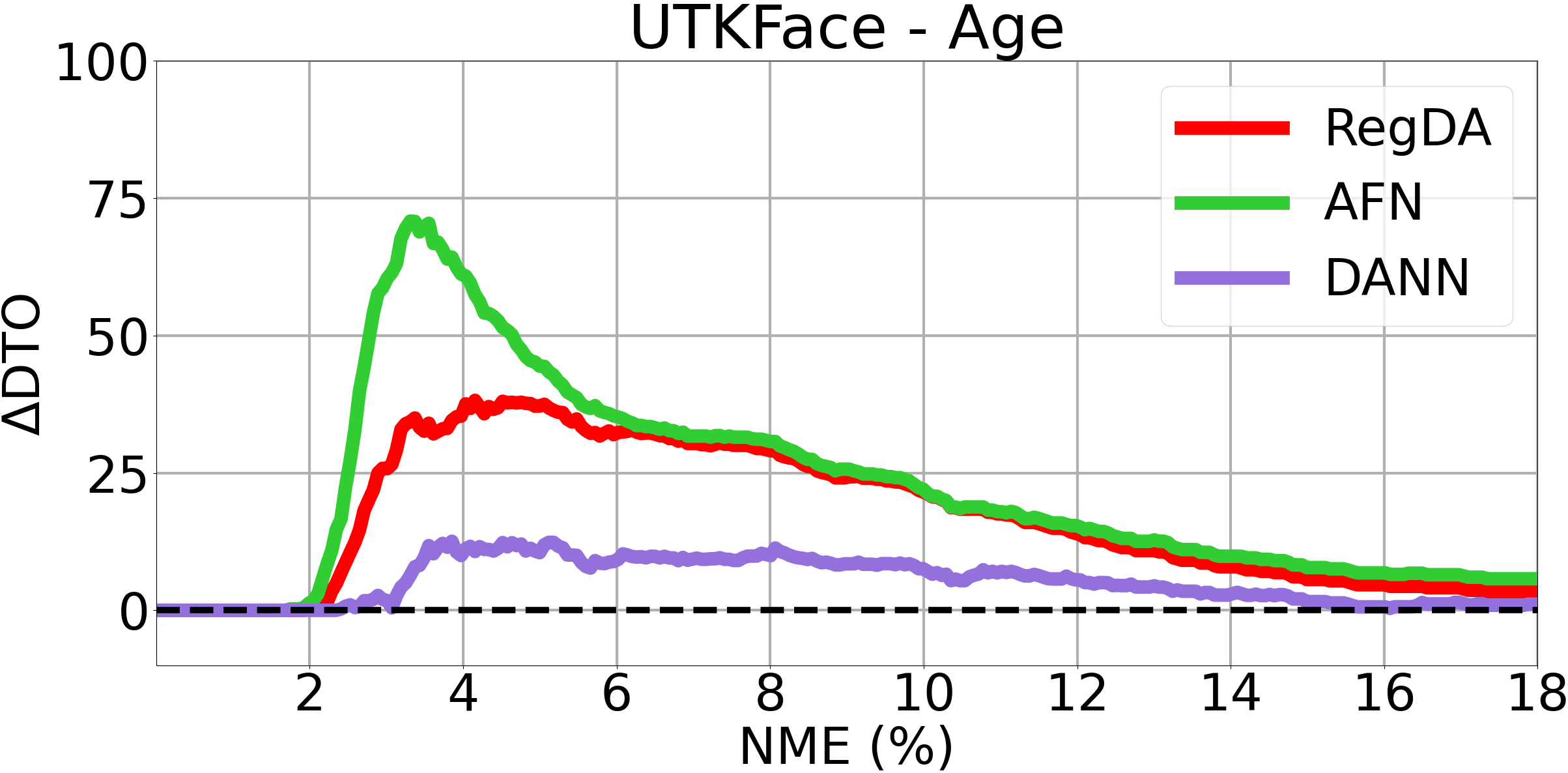}\\
\end{tabular}
\caption{Landmark detection results. Comparison among the cross-domain methods and the reference baseline in terms of $HF$, $\sigma(HF)$ and $\Delta DTO$ when changing the NME threshold used for $SDR$.
}
\label{fig:NMEth_app} \vspace{-3mm}
\end{figure}

\begin{table*}[t]
    \begin{center}
    \small
    \resizebox{\textwidth}{!}{
    \begin{NiceTabular}{|c | c c c c | c c | c | c || c c c c | c c | c | c |}[colortbl-like,cell-space-limits=1.5pt]
        \hline
        {} & \multicolumn{16}{c}{\textbf{CelebA - EyeBags}}\\
        \hline
        {} & \multicolumn{8}{c||}{{\textit{Male/Female} $\longrightarrow$ \textit{Young/Old}}} & \multicolumn{8}{c|}{{\textit{Young/Old} $\longrightarrow$ \textit{Male/Female}}} \\
        \cline{2-17}
        {} & Acc. & MGA & mGA & DA & \textcolor{gray}{DEO} & \textcolor{gray}{DEOdds} & $\Delta DTO$ & $\mathbf{\sigma(HF)}$ & Acc. & MGA & mGA & DA & \textcolor{gray}{DEO} & \textcolor{gray}{DEOdds} & $\Delta DTO$ & $\mathbf{\sigma(HF)}$ \\ \hline
        
Baseline \cite{wang2020towards} & 81.06	& 83.20 & 74.05 & 9.16 & \textcolor{gray}{11.54} &	\textcolor{gray}{24.69} & 0.00 & 0.500 $\pm$ 0.000 
& 82.55 & 89.17 & 72.04 & 17.13 & \textcolor{gray}{19.72} & \textcolor{gray}{41.95} & 0.00 & 0.500 $\pm$ 0.000 \\ 
\hline

\textbf{AFN} \cite{xu2019larger} & 81.31	& 83.30 & 74.92 & \textbf{8.38} & \textcolor{gray}{9.06} &	\textcolor{gray}{\underline{21.42}} & 0.78 & 0.554 $\pm$ 0.014
& 81.35 & 88.64 & 69.77 & 18.87 & \textcolor{gray}{23.71} & \textcolor{gray}{50.51} & -2.31 & 0.361 $\pm$ 0.068 \\ 

\textbf{DANN} \cite{ganin2016domain}  & \textbf{83.48} &	\textbf{85.83} &	\textbf{76.16} &	9.67 & \textcolor{gray}{18.28} &	\textcolor{gray}{31.74} & \textbf{3.18} & \textbf{0.626} $\pm$ 0.039 
& \textbf{82.64} & \textbf{89.50} & \textbf{71.75} & \underline{17.75} & \textcolor{gray}{21.83} & \textcolor{gray}{42.73} & \textbf{-0.15} & \textbf{0.481} $\pm$ 0.021 \\ 

\hline
\textbf{GroupDRO} \cite{sagawa2019distributionally} & \underline{82.52} &	\underline{84.90} &	\underline{75.68} &	\underline{9.22}	& \textcolor{gray}{15.48} &	\textcolor{gray}{29.15} & \underline{2.29} & \underline{0.599} $\pm$ 0.055 
& 82.10 & 89.36 & 70.58 & 18.79 & \textcolor{gray}{32.38} & \textcolor{gray}{56.68} & -1.30 & 0.408 $\pm$ 0.098 \\ 

\textbf{g-SMOTE} \cite{zietlow2022leveling}  & 80.16 &	82.58 &	72.62 &	9.96  & \textcolor{gray}{\underline{8.89}} &	\textcolor{gray}{22.18} & -1.54 & 0.415 $\pm$ 0.168 
& 82.01 & 88.77 & \underline{71.38} & \textbf{17.34} & \textcolor{gray}{\underline{15.66}} & \textcolor{gray}{\underline{36.10}} & -0.76 & \underline{0.458} $\pm$ 0.132 \\ 

\textbf{FSCL} \cite{park2022fair}  & 80.45 &	84.65 &	69.35 &	15.30 & \textcolor{gray}{\textbf{8.64}} &	\textcolor{gray}{\textbf{14.38}} & -3.37 & 0.235 $\pm$ 0.097 
& \underline{82.32} & \underline{89.30} & 71.25 & 18.05 & \textcolor{gray}{\textbf{11.73}} & \textcolor{gray}{\textbf{28.31}} & \underline{-0.69} & 0.450 $\pm$ 0.029 \\ 

\hline

\rowcolor[gray]{0.95}
         & \multicolumn{8}{c||}{\textit{Young/Old} $\longrightarrow$ \textit{Young/Old} (\textbf{Oracle})} & \multicolumn{8}{c|}{\textit{Male/Female} $\longrightarrow$ \textit{Male/Female} (\textbf{Oracle})}\\
        \hline
        
\rowcolor[gray]{0.95}
\textbf{AFN} \cite{xu2019larger} & 81.70	& 83.81 & 75.04 & 8.77 & \textcolor{gray}{10.92} &	\textcolor{gray}{23.43} & 1.16 & 0.561 $\pm$ 0.013
& 83.59 & 89.34 & 74.47 & 14.87 & \textcolor{gray}{3.55} & \textcolor{gray}{12.05} & 2.32  & 0.646 $\pm$ 0.032 \\ 

\rowcolor[gray]{0.95}
\textbf{DANN} \cite{ganin2016domain}  & 83.00 &	84.90 &	77.01 &	7.89 & \textcolor{gray}{7.42} &	\textcolor{gray}{16.74} & 3.41  & 0.676 $\pm$ 0.040
& 83.82 & 90.28 & 73.56 & 16.72 & \textcolor{gray}{33.71} & \textcolor{gray}{48.89} & 1.81  & 0.594 $\pm$ 0.049 \\ 
\hline

\rowcolor[gray]{0.95}
\textbf{GroupDRO}  \cite{sagawa2019distributionally}  & 82.23 & 83.70 & 75.67 & 8.03 & \textcolor{gray}{16.99} &	\textcolor{gray}{35.92} & 1.63 & 0.598 $\pm$ 0.049
& 83.08 & 89.25 & 73.28 & 15.98 & \textcolor{gray}{16.30} & \textcolor{gray}{30.82} & 1.18 & 0.575 $\pm$ 0.075 \\ 

\rowcolor[gray]{0.95}
\textbf{g-SMOTE} \cite{zietlow2022leveling}  & 	81.21 &	81.99 &	73.71 &	8.28 & \textcolor{gray}{6.63} &	\textcolor{gray}{17.97} & -0.95 & 0.477 $\pm$ 0.072
& 82.00 & 88.94 & 72.38 & 16.56 & \textcolor{gray}{28.11} & \textcolor{gray}{46.63} & 0.23 & 0.521 $\pm$ 0.013 \\ 

\rowcolor[gray]{0.95}
\textbf{FSCL} \cite{park2022fair}  & 80.50 &	84.66 &	69.39 &	15.27 & \textcolor{gray}{8.72} &	\textcolor{gray}{14.44} & -3.33 & 0.237 $\pm$ 0.098
& 82.89 & 89.56 & 72.29 & 17.27 & \textcolor{gray}{34.51} & \textcolor{gray}{46.02} & 0.37 & 0.515 $\pm$ 0.023 \\ 

 \hline
        
    \end{NiceTabular}
    }
    \end{center}
    \caption{Model Transferability analysis on the classification task. All the relative metrics are calculated with respect to the baseline results in the first row. 
    }
    \label{tab:transfer}\vspace{-2mm}
\end{table*}

\begin{table*}[t] 
    \begin{center}
    \small
    \resizebox{\textwidth}{!}{
    \begin{NiceTabular}{| c | c c c c | c | c || c c c c | c | c |}[colortbl-like,cell-space-limits=1.5pt]
        \hline
        {} & \multicolumn{12}{c}{\textbf{UTK Face - Landmark Detection}}\\
        \hline
        {} & \multicolumn{6}{c|}{{\textit{Skin Tone} $\longrightarrow$ \textit{Age}}} 
        & \multicolumn{6}{c|}{{\textit{Age} $\longrightarrow$ \textit{Skin Tone}}} 
        \\
        \cline{2-13}
        
        {} & SDR & MGS & mGS & DS & $\Delta DTO$ & $\mathbf{\sigma(HF)}$
        &  SDR & MGS & mGS & DS & $\Delta DTO$ &  $\mathbf{\sigma(HF)}$ 
\\ \hline
        
Baseline & 88.30 & 90.14  & 85.13 & 5.01 & 0.00 &  0.500 $\pm$ 0.000 
& 82.35  & 86.67 & 77.97 & 8.70 & 0.00 & 0.500 $\pm$ 0.000 
\\ \hline

\textbf{AFN} \cite{xu2019larger} & \textbf{91.46} & \textbf{93.08} & \textbf{89.84} & \underline{3.23} & \textbf{5.55} & \textbf{0.760} $\pm$  0.116  
& \underline{88.03} & \underline{91.67} & \underline{84.33} & \underline{7.34} & \underline{8.00} & \underline{0.828} $\pm$ 0.152 
\\ 

\textbf{DANN} \cite{ganin2016domain} & 79.55 & 80.91 & 78.07 & \textbf{2.85} & -11.23 & 0.126 $\pm$ 0.074 
& 79.83 & 80.84 & 78.82 & \textbf{2.02} & -2.81 & 0.501 $\pm$ 0.033 
\\ 

\textbf{RegDA} \cite{jiang2021regda} & \underline{89.64}  & \underline{91.82} & \underline{87.64} & 4.18 & \underline{3.02} & \underline{0.650} $\pm$ \underline{0.092} 
& \textbf{90.55} & \textbf{94.59} & \textbf{86.44} & 8.15 & \textbf{11.15} & \textbf{0.883} $\pm$ 0.040
\\ 

\hline

\rowcolor[gray]{0.95}
         {} & \multicolumn{6}{c|}{\textit{Age} $\longrightarrow$ \textit{Age} (\textbf{Oracle})} 
         & \multicolumn{6}{c|}{\textit{Skin Tone} $\longrightarrow$ \textit{Skin Tone} (\textbf{Oracle})}
         \\
        \hline

\rowcolor[gray]{0.95}
\textbf{AFN} \cite{xu2019larger} & 91.22 & 92.03 & 90.49 & 1.55 &5.43 &  0.792 $\pm$ 0.009 
& 91.32 & 91.94 & 90.68 & 1.27 & 13.43 & 0.960 $\pm$ 0.020 
\\ 

\rowcolor[gray]{0.95}
\textbf{DANN} \cite{ganin2016domain} & 80.10 & 80.77 & 79.45 & 1.32 & -10.30 &  0.160 $\pm$ 0.099 
& 83.20 & 86.03  & 78.85 & 7.18 & 0.40 & 0.551 $\pm$ 0.060
\\ 

\rowcolor[gray]{0.95}
\textbf{RegDA} \cite{jiang2021regda} & 91.28 & 91.61 & 90.61 & 1.00  & 5.25 & 0.796 $\pm$ 0.043
& 92.69 & 94.78 & 89.88 & 4.90 & 14.36 & 0.950 $\pm$ 0.014 
\\ 

\hline 

    \end{NiceTabular}
    }
    \end{center} 
    \caption{Model transferability analysis on the landmark detection task.
    }
    \label{tab:transfer_landmark}
    \vspace{-4mm}
\end{table*}

We present here for completeness the results of our analysis for the classification problem on the set of 13 attributes already used in \cite{ramaswamy2021fair,zietlow2022leveling}. According to \cite{ramaswamy2021fair}, these attributes are the most reliable out of the whole set of 40 CelebA attributes as they can be labeled objectively, without being ambiguous for a human.

Table \ref{tab:celeba13_5} shows the results on {CelebA} for the binary facial attribute prediction task. Specifically, each number corresponds to the average over 13 experiments done considering every binary attribute alone (\ie 13 different binary classifiers). 
The cross-domain models are presented in the different horizontal sections, sorted by family: Regularization, Adversarial, Feature Alignment, and Self-Supervised-based techniques. In all the cases we managed the gender sensitive attributes as domain. 
The bottom part of the table contains the SOTA fairness approaches. 

Made exceptions for SWAD, Fish, and FSCL, all the approaches exceed the baseline (\ie ${\sigma(HF)} > 0.500$).  The best approach is AFN which is able to increase both $mGA$ and $MGA$, decreasing $DA$. The second best is DANN confirming the effectiveness of adversarial techniques to deal with the fairness problem \cite{wang2020towards,wang2022fairness}. Considering the high baseline accuracy, the improvements of the different methods appear relatively small but they are consistent with the results presented in \cite{zietlow2022leveling} (Supplementary Table S1).

As a final note, we share our initial surprise in collecting very low results for FSCL. This approach is based on contrastive learning and is designed to avoid encoding sensitive information in the learned embedding space, thus it is very much in line with the cross-domain logic. However as any contrastive learning approach, its limitation is in the amount of data needed to fully train the model which makes it ineffective in our experimental setting.

\subsection{Landmark Detection - Further Analysis}

In the interest of comprehensiveness, we integrate the curves already presented in Figure 4 of the main submission by including here in Figure \ref{fig:NMEth_app} the results on both \emph{Skin Tone} and \emph{Age}. For the former the behavior of the studied methods is consistent in showing an advantage over the baseline regardless of the used metric.
This is not the case for the latter, as already discussed in the main paper. We also include the curves of ${\sigma(HF)}$ that show the same trend of $HF$ in terms of model ranking. 

Although no previous publication proposed an unfairness mitigation approach for landmark detection, GroupDRO might sound general enough to be applied also in this setting. This approach dynamically adjusts loss weights during optimization to prioritize the poorest-performing protected group. However, our investigation revealed that, even after a comprehensive hyperparameter search, the loss of the worst group decreases extremely slowly in landmark detection, and the method keeps focusing on it which ultimately makes it unable to obtain an improvement neither on the best group nor overall. The result is a high  Normalized Mean Error (NME) achieved by  GroupDRO during training and a consequent 0\% Success Detection Rate (SDR) on the test set. Notably, applying looser thresholds did not improve the situation, suggesting that the logic employed by GroupDRO may not be well-suited for landmark detection tasks.

\subsection{Model Transferability - Further Analysis}
We present here further results of the experiments performed to study the transferability of unfairness mitigation models. Specifically for the classification task, we consider \emph{age} as the initial sensitive attribute and we test the obtained model to evaluate whether it helps in mitigating unfairness with respect to \emph{gender}. From the results we observe that the advantage obtained with the gender-robust model on different age groups is not symmetric: the results in the right part of Table \ref{tab:transfer} show that none of the methods improve over the baseline. 
The transferability results look instead always very promising on landmark detection where a model trained to be fair on \emph{skin tone} is effective also in reducing the performance gap among different \emph{age} groups and vice-versa as shown in Table \ref{tab:transfer_landmark}.
Overall the possibility to reuse fair models on different sensitive attributes connects with the ability of the models to capture knowledge shared across them and generalize at deployment time to new social conditions with different ethical constraints. We find it an interesting aspect that gives rise to new research questions and deserves more attention in the future.

\end{document}